%% file: main.tex
\documentclass{article}

\usepackage[preprint]{corl_2026} 

\usepackage{amsmath}
\usepackage{amssymb}
\usepackage{booktabs}
\usepackage{graphicx}
\usepackage{makecell}
\usepackage{multirow}
\usepackage{array}
\usepackage{wrapfig}

\usepackage{caption}
\usepackage{subcaption}

\usepackage{booktabs}
\usepackage{xcolor}
\usepackage{pifont}

\newcommand{\cmark}{\textcolor{green!70!black}{\ding{51}}}
\newcommand{\xmark}{\textcolor{red}{\ding{55}}}

\DeclareRobustCommand{\ScoutVLALogo}{\raisebox{-0.16em}{\includegraphics[height=1.15em]{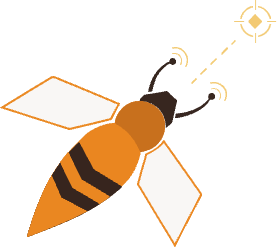}}}

\title{\ScoutVLALogo\ ScoutVLA: UAV-Centric Active Perception via a Dual-Expert VLA Model for Open-World Embodied Question Answering}

\author{
  Wenhao Lu$^{1}$, Zhengqiu Zhu$^{1}$, Xiaofeng Wang$^{2}$, \\
  \textbf{Xiaoran Zhang$^{1}$}, \textbf{Yatai Ji$^{1}$}, \textbf{Yong Zhao$^{1}$}, \\
  \textbf{Yue Hu$^{1}$}, \textbf{Yingzhen Nie$^{1}$}, \textbf{Jinlong Zhu$^{1}$}, \textbf{Zheng Zhu$^{2}$} \\[1.5ex]
  $^{1}$National Key Laboratory of Digital Intelligent Modeling and Simulation, \\
  National University of Defense Technology \\
  $^{2}$GigaAI \\[1ex]
  \texttt{\{luwenhao20, zhuzhengqiu12\}@nudt.edu.cn}
}

\begin{document}
\maketitle

\begin{abstract}
Aerial Embodied Question Answering (EQA) requires Unmanned Aerial Vehicles (UAVs) to actively perceive the environment and answer natural language questions. Existing outdoor EQA systems usually stop once the target enters the UAV's field of view, leaving the fine-grained viewpoint adjustment needed for evidence-seeking questions largely unresolved. To address this issue, we introduce FG-EQA, a fine-grained active perception EQA benchmark with more than 40K simulated trajectories and 1K real-world trajectories. Drawing inspiration from the ``waggle dance'' of scout bees, which iteratively adjust their flight paths to verify target information, we propose ScoutVLA, an evidence-driven Vision-Language-Action model for outdoor EQA. 
To emulate this active exploration behavior, ScoutVLA features a decoupled dual-expert architecture: a vision-language expert infers the semantic intent to identify missing evidence, while an independent action expert employs high-DoF flow matching to generate continuous viewpoint-refinement trajectories. 
To balance the competing demands of continuous control and semantic reasoning, we devise a decoupled training strategy with a knowledge insulation mechanism that prevents the action gradients from erasing the model's multimodal reasoning ability.
Extensive simulated experiments and a qualitative real-world field study both verify the superiority of ScoutVLA over the state-of-the-art baselines, demonstrating a \textbf{10.48$\boldsymbol{\times}$} higher average strict success rate and a \textbf{7.72$\boldsymbol{\times}$} higher average QA correctness.
\end{abstract}

\keywords{Embodied Question Answering, Active Perception, Bio-inspired, Vision-Language-Action Model}

\section{Introduction}
\label{sec:introduction}

Unmanned Aerial Vehicles (UAVs) have emerged as flexible mobile sensors for urban monitoring and scene understanding. A compelling advancement in this domain is aerial Embodied Question Answering (EQA), where a UAV explores an environment to gather visual evidence for answering natural language questions. However, current outdoor EQA pipelines typically adopt a decoupled navigate-then-answer paradigm~\citep{zhao2025cityeqa,varghese2025bridgeeqa}, in which a UAV agent halts when the target enters the field of view or is in close proximity. Unfortunately, this premature termination leaves the UAV at poor viewing angles and distances, obscuring key visual evidence needed for accurate QA generation.

\begin{wrapfigure}{r}{0.7\textwidth}
    \vspace{-1em} 
    \centering
    \includegraphics[width=\linewidth]{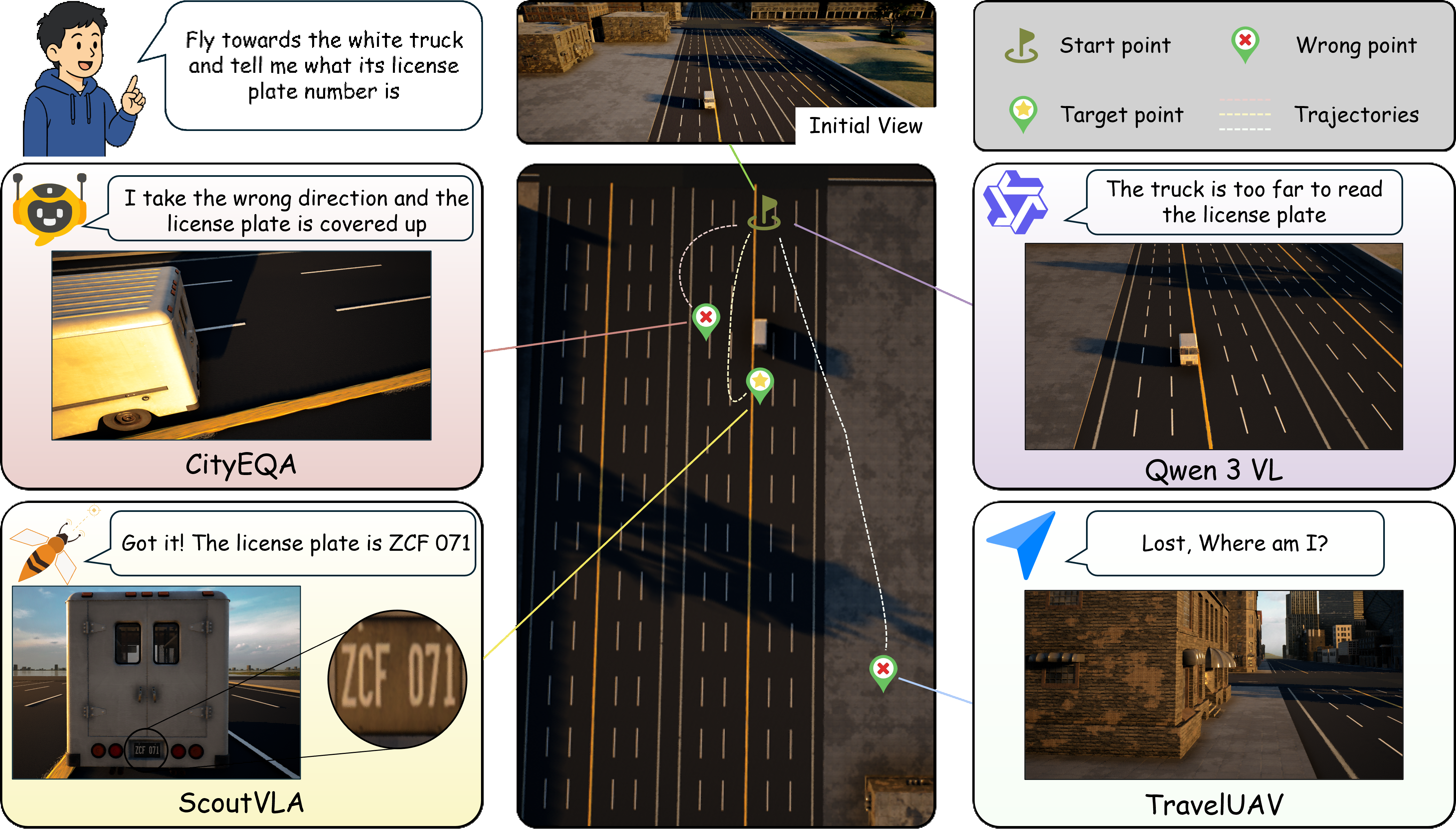}
    \caption{\textbf{Task illustration of question-conditioned fine-grained active perception.}}
    \label{fig:problem}
    \vspace{-1em} 
\end{wrapfigure}

The inherent limitations of this paradigm are illustrated in Figure~\ref{fig:problem}. In the initial view, a white truck is already clearly visible on the road. When tasked to read its license plate, standard baselines falter in varied ways: a passive pure vision-language model (e.g., Qwen3-VL) struggles due to the extreme distance, while a conventional EQA agent (e.g., CityEQA) might successfully fly closer but arbitrarily halt alongside the truck's flank, completely missing the rear-mounted plate. Moreover, the navigation model (e.g., TravelUAV) gets lost and completely deviates from the target direction. Acquiring precise, localized evidence requires the UAV to interpret the question's semantic intent and execute continuous spatial maneuvers to align with specific locations like the rear bumper. Recognizing this critical deficit in current studies, we formalize this neglected late-stage EQA phase as \emph{question-conditioned fine-grained active perception}.


Addressing this active perception task presents three formidable challenges. \textbf{First, open-world visual evidence is exceptionally viewpoint-sensitive.} A fractional deviation in spatial position and camera pose dictates whether a tiny, angle-dependent feature is legible. \textbf{Second, establishing a semantic-to-physical mapping is inherently difficult.} The agent must bridge a vast cross-modal gap, translating highly abstract linguistic intents directly into concrete 3D geometric search strategies. \textbf{Third, high-DoF continuous control is difficult.} Active perception requires continuous adjustment of position and pose to align with obscured target regions, while existing methods often rely on fixed multi-view camera arrays or low-dimensional discrete action spaces, thus failing to support the precise viewpoint refinement required for reliable evidence acquisition.

To systematically address this problem, we introduce \textbf{FG-EQA}, the first benchmark dedicated to fine-grained perception in aerial EQA. Unlike long-horizon navigation datasets, each episode in FG-EQA isolates the evidence-seeking phase: the target is visible at the onset, yet necessitates a complex sequence of active viewpoint refinements to accurately answer the query. Specifically, it includes over 40K simulated trajectories and 1K human-piloted real-world trajectories.

Furthermore, inspired by the ``waggle dance'' of scout bees, which iteratively hover and refine their flight paths to confirm discoveries, we propose \textbf{ScoutVLA}, a powerful dual-expert Vision-Language-Action (VLA) model trained with a decoupled strategy. To tackle the semantic-to-physical mapping, a PaliGemma-based multimodal expert perfectly aligns the question semantics with the missing visual evidence. To overcome viewpoint-sensitivity and continuous-control hurdles, an independent action expert employs high-DoF flow matching to generate continuous viewpoint-refinement control. Finally, to prevent continuous action regression from catastrophically overriding discrete language reasoning, we devise a two-stage training paradigm. By integrating knowledge insulation with parameter-efficient LoRA and LMAdapter modules, we successfully orchestrate evidence-grounded QA without inducing catastrophic forgetting of continuous navigation capabilities.

\textbf{Contributions.} We formalize the problem of question-conditioned fine-grained active perception and introduce the FG-EQA benchmark, comprising over 40K simulated and 1K real-world trajectories. To tackle this challenge, we propose ScoutVLA, a bio-inspired dual-expert VLA model that translates semantic intent into continuous 5-DoF viewpoint-refinement maneuvers. Moreover, we establish a decoupled two-stage training paradigm that harmonizes the conflicting demands of spatial control and discrete reasoning text generation. Extensive sim-to-real experiments on FG-EQA verify the superiority of ScoutVLA over state-of-the-art baselines, demonstrating a 10.48$\times$ higher average strict success rate and a 7.72$\times$ higher average QA correctness.

\section{Related Work}
\label{sec:related}

\textbf{Embodied question answering.} Early EQA focused on indoor navigation coupled with QA~\citep{das2017embodiedqa,majumdar2024openeqa,cheng2025efficienteqa,zhai2025memory,jiang2025beyond}. As studies transitioned to outdoor environments~\citep{zhao2025cityeqa,varghese2025bridgeeqa,li2025industryeqa}, they adopted modular navigate-then-answer pipelines. However, these systems implicitly assume that reaching a target's general vicinity resolves the query, thereby entirely overlooking the fine-grained viewpoint adjustment necessary for angle-dependent visual evidence. ScoutVLA specifically targets this missing active perception phase. 

\textbf{Active perception and aerial VLA models.} Recent multi-modal autonomous agents~\citep{hu2026longnav,zha2025aircopbench,he2026activezero} and aerial VLA models~\citep{gao2025openfly,cai2025flightgpt,wang2024traveluav,li2025skyvln,wang2025uavflow,wu2025vlaan} have successfully mapped language instructions to flight control. Yet, despite their active mobility, they primarily execute macroscopic routing (e.g., ``fly to the building'') utilizing low-DoF or discrete action spaces. ScoutVLA advances this paradigm by integrating the linguistic query directly into continuous 5-DoF control to actively reveal occluded evidence. \textbf{A comprehensive discussion of related literature, alongside an architectural comparison table of recent VLA models, is provided in Appendix~\ref{app:related_vla}.}

\section{FG-EQA Benchmark}
\label{sec:dataset}

\subsection{Task Formulation}
\label{subsec:task}
Each episode starts from an initial UAV view in which the target is visible, but the available visual cues are insufficient to correctly answer the question. At time $t$, the UAV receives an RGB perception $I_t$, state $\mathbf{s}_t$, and question $q$, executes a viewpoint-refinement command $\mathbf{u}_t$, and finally produces a short answer $y$. The goal is not merely to approach the target, but to induce a trajectory $\tau$ that exposes the requisite visual evidence. Formally, as defined in Equation~\ref{eq:objective}, we aim to find the optimal policy parameters $\theta^*$ that maximize the expected log-likelihood of generating the correct answer. Here $y^*$ is the ground-truth short answer and $\tau$ is the trajectory induced by policy $\pi_{\theta}$. This formulation couples viewpoint selection with answer generation: success depends on whether the policy acquires question-relevant evidence, not only on terminal distance to the target.

\begin{equation}
    \theta^* = \arg\max_{\theta} \mathbb{E}_{\tau \sim \pi_{\theta}}
    \left[\log p_{\theta}(y^* \mid \tau, q)\right].
    \label{eq:objective}
\end{equation}

For fine-grained perception, the UAV acts in a 5-DoF viewpoint-refinement space, $\mathbf{u}_t=[\Delta x_t,\Delta y_t,\Delta z_t,\Delta \psi_t,\Delta \theta^g_t]$, where the first three dimensions translate the UAV and the last two adjust body yaw and gimbal pitch. This action space is designed for local evidence acquisition after the target has appeared, rather than long-horizon target search. State and model-prediction action encodings are provided in Appendix~\ref{app:dataset}.

\subsection{Dataset Construction}
\label{subsec:data_construction}

To bridge the gap between macroscopic spatial reasoning and fine-grained visual evidence acquisition, we construct the FG-EQA dataset through a systematic two-phase pipeline: (1) Perception Trajectory Generation, which formulates viewpoint-refinement flight paths; and (2) QA Data Annotation, which aligns terminal evidence with complex natural language queries.

\textbf{Perception Trajectory Generation.} 
We initiate episodes by reusing simulated outdoor scenes from UAV-ON~\citep{xiao2025uavon}. To construct varied starting poses mimicking challenging initial views, we apply yaw offsets ($\pm15^\circ$, $\pm30^\circ$) from the target-oriented optical axis, utilizing a Vision Language Model (VLM) to discard poses where the target is entirely obscured. Next, to define ideal observation endpoints, we uniformly sample candidate views on a target-centered continuous hemispherical grid. These candidates undergo rigorous filtering by a panel of UAV experts to select optimal terminal views ensuring structural completeness, visual clarity, and semantic richness of the target. Finally, base trajectories are collected by navigating from initial positions to these selected endpoints while keeping the target centered. To endow ScoutVLA with dynamic error-correction capabilities, we inject random pose perturbations into 30\% of the trajectories to act as negative recovery samples. To evaluate model physical transferability, we also collect 1K real-world trajectories. Unlike the automated simulation pipeline, these physical episodes are captured by professional UAV pilots manually executing analogous micro-viewpoint refinement maneuvers around real-world targets, recording unconstrained onboard camera streams from initial discovery to final evidence acquisition.

\textbf{QA Data Annotation.} 
Based on the validated endpoint views, we define seven critical outdoor question types. An initially prompted VLM generates candidate reasoning chains and short-answer pairs based on the terminal view. To guarantee that the benchmark truly assesses active perception, we apply a stringent two-round filtering mechanism: an initial VLM coarse filter eliminates factual errors and questions that are prematurely answerable from the \emph{initial} view, followed by strict human verification for QA validity. To further mitigate overfitting to simulated textures and enhance visual-language generalizability, we introduce an additional 80K generic VQA samples from LLaVA-Instruct-150K~\citep{liu2023llava}. Together, this composite dataset of over 120K examples robustly grounds the visual-language reasoning capabilities required for fine-grained outdoor EQA. Comprehensive details regarding the automated evaluation pipeline, data collection prompts, and benchmark statistics are provided in Appendix~\ref{app:dataset}.

\section{ScoutVLA}
\label{sec:model}

To systematically address the challenges of active perception mentioned in Section~\ref{sec:introduction}---viewpoint sensitivity, semantic-to-physical mapping, and high-DoF continuous control---we model ScoutVLA as a dual-expert end-to-end architecture. This model consists of two key modules: a PaliGemma~2-based multimodal expert~\citep{steiner2024paligemma2}, and an independent action expert network. Furthermore, we introduce a decoupled two-stage training strategy to ensure that the UAV agent establishes robust flight control capabilities while successfully retaining visual question-answering skills. The comprehensive system overview is illustrated in Figure~\ref{fig:model}. Full implementation details and hyperparameter configurations are provided in Appendix~\ref{app:model} and Appendix~\ref{app:training}.

\begin{figure}[t]
    \centering
    \includegraphics[width=\linewidth]{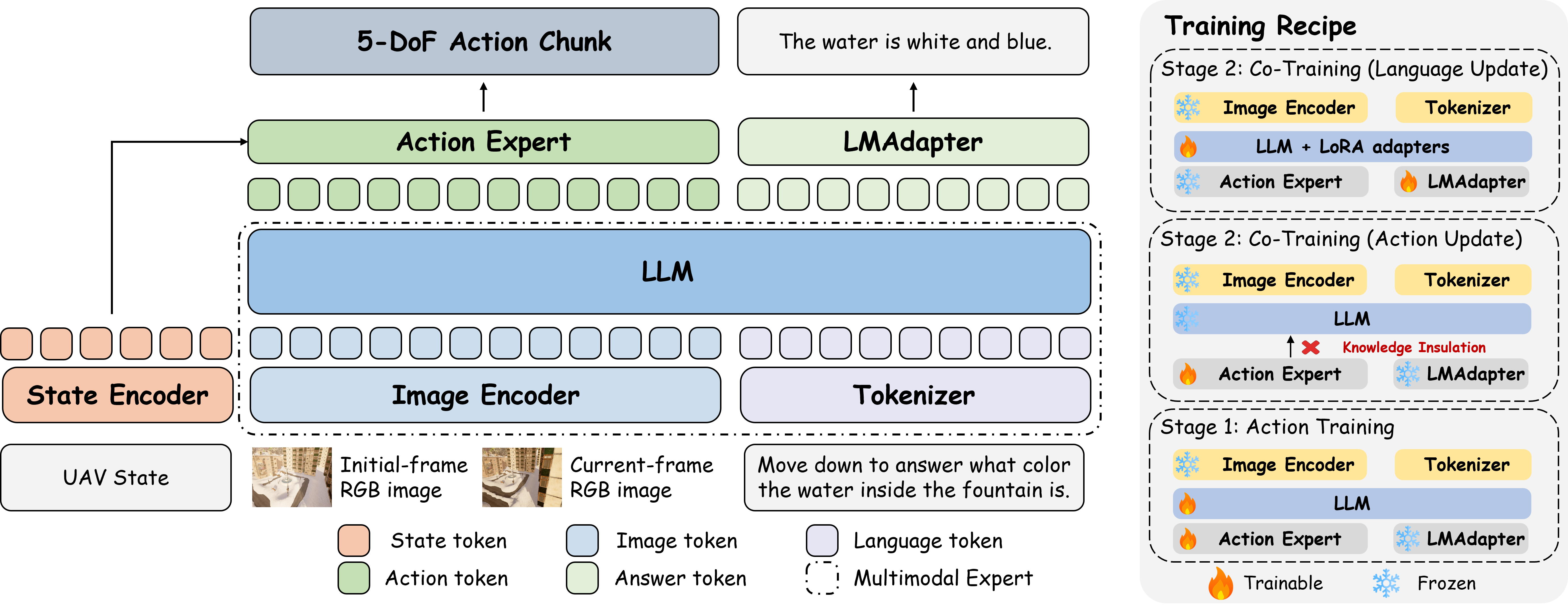}
    \caption{\textbf{ScoutVLA overview.} Left: dual-expert architecture for question-conditioned viewpoint control and short-answer generation. Right: two-stage training recipe.}
    \label{fig:model}
    \vspace{-1.5em}
\end{figure}

\subsection{Architecture}
\label{subsec:architecture}

ScoutVLA abandons explicit modular boundaries to process unified inputs and produce continuous navigational trajectories natively coupled with textual answers. 

The first core component is a PaliGemma~2-based multimodal expert, which is fundamentally responsible for visual encoding and semantic reasoning. Visual tokens extracted by SigLIP~\citep{zhai2023siglip} are contextually fused with language inputs through bidirectional attention. This process effectively bridges the semantic-to-physical gap, enabling the network to infer what answers can be derived from the current view and what target details remain hidden to guide the next viewpoint adjustment.

The second core component is an independent action expert for high-DoF trajectory generation. Sharing the cross-modal context representations provided by the multimodal expert, the action expert utilizes flow matching~\citep{lipman2022flowmatching} to smoothly synthesize continuous 5-DoF flight maneuvers. This design bypasses discrete action quantization, directly targeting the fine-grained spatial maneuvers required to overcome viewpoint sensitivity.

Finally, to harmonize the severe paradigm discrepancies between generating real-valued action fields and outputting discrete text tokens, we introduce tailored structural adaptations at the language output side. Rather than naively mixing the gradients into a shared head, we place an LMAdapter module independently after the multimodal backbone. This design supports continuous flow-matching regression while preserving the ability to generate autoregressive natural language text, which motivates the two-stage training below.

\subsection{Two-Stage Training Recipe}
\label{subsec:training}


Although the structural adaptations introduced above accommodate divergent output formats, jointly training both capabilities from scratch remains fundamentally unstable. When the pretrained multimodal backbone is updated predominantly by the flow matching regression loss, its hidden representations become biased toward control-oriented action denoising, thereby weakening the visual-text alignment required for subsequent answer generation. Conversely, naively mixing flow matching with autoregressive language modeling introduces gradients with drastically different scales, causing action and language updates to brutally interfere with each other~\citep{zhou2025chatvla,zhou2025chatvla2}. Consequently, ScoutVLA implements a decoupled two-stage training paradigm.

\textbf{Stage 1: action-only control learning.}
Given a clean action chunk $\mathbf{u}$, Gaussian noise $\boldsymbol{\epsilon}$, interpolation scalar $\lambda$, interpolated sample $\mathbf{x}_{\lambda}=\lambda\boldsymbol{\epsilon}+(1-\lambda)\mathbf{u}$, and multimodal condition $\mathbf{c}$, Stage~1 optimizes a continuous vector field objective to minimize the action denoising error, as shown in Equation~\ref{eq:stage1_loss}. During this stage, all trainable model parameters are updated to establish the visual-action mapping.

\begin{equation}
    \mathcal{L}_{\mathrm{stage1}}
    =
    \mathbb{E}_{\lambda,\mathbf{u},\boldsymbol{\epsilon}}
    \left[
    \left\|
    \mathbf{v}_{\theta}(\mathbf{x}_{\lambda},\lambda,\mathbf{c})
    -(\boldsymbol{\epsilon}-\mathbf{u})
    \right\|_2^2
    \right].
    \label{eq:stage1_loss}
\end{equation}
\textbf{Stage~2: action-language joint training with knowledge insulation.} Stage~2 introduces language task data to activate the specific EQA capabilities. Given the terminal first person view (FPV) perception $I_T$, the language objective is a short-answer next-token loss which is shown in Equation~\ref{eq:language_loss}.
\begin{equation}
    \mathcal{L}_{\mathrm{stage2}} =
    \begin{cases}
        \mathcal{L}_{\mathrm{action}}, & \text{action step},\\
        \mathcal{L}_{\mathrm{language}}, & \text{language step}.
    \end{cases}
    \label{eq:stage2_loss}
\end{equation} 

\begin{equation}
    \mathcal{L}_{\mathrm{language}}
    =
    -\sum_{j=1}^{|y|-1}
    w_j \log p_{\theta}(y_{j+1}\mid I_T,q,y_{1:j}).
    \label{eq:language_loss}
\end{equation}
\section{Experiments}
\label{sec:experiments}
To evaluate ScoutVLA, we design comprehensive simulated and real-world experiments to answer four research questions: 
\textbf{(RQ1)} Does ScoutVLA outperform the state-of-the-art baselines on the active perception task? 
\textbf{(RQ2)} How do the two-stage training paradigm and specific language-recovery modules contribute? 
\textbf{(RQ3)} Does the active perception process yield more answerable evidence at final viewpoint? 
\textbf{(RQ4)} Behaviorally, how does ScoutVLA transfer to real-world scenarios? Also, Appendix~\ref{app:experiments_deployment} discusses its success cases alongside its failure modes.

\subsection{Setup}
\label{subsec:setup}

\textbf{Environments and Metrics.} All simulation experiments are conducted in nine high-fidelity environments, and all test targets are unseen during training. We report navigation quality with Target Detection Rate (TDR), terminal Distance to Target (DT), and Average Trajectory Length (ATL); answer quality is evaluated with LLM-judged QA Score (QAS) and QA Correctness (QAC); and overarching end-to-end performance is heavily weighted by the Strict Success Rate (SSR). To systematically mitigate evaluator variance, three independent LLM judges score each generated answer, and we average their ratings following established protocols in OpenEQA~\citep{majumdar2024openeqa} and CityEQA~\citep{zhao2025cityeqa}. 

\textbf{Implementation Details.} The exhaustive hyperparameter configurations for both training stages are provided in Appendix~\ref{app:evaluation}.

\subsection{Baselines}
\label{subsec:baselines}

We compare ScoutVLA with baselines covering seven evaluation families. \textbf{Random} executes randomly sampled navigation actions before final-frame QA. \textbf{Passive QA} includes blind text-only LLM baselines, which receive the question without visual input to test language priors, and an initial-view VQA baseline, which answers directly from the start frame. \textbf{General VLM} evaluate whether general multimodal models can act as zero-shot perception agents. \textbf{Navigation} baselines, including OpenFly~\citep{gao2025openfly} and TravelUAV~\citep{wang2024traveluav}, are designed for UAV navigation rather than answer generation. \textbf{Aerial VLA} uses UAV-Flow~\citep{wang2025uavflow}, which generates UAV actions but has no native QA head, to test whether continuous UAV action generation alone is sufficient. Policy-only baselines without native answer generation are evaluated by using Qwen3-VL to answer from their terminal FPV image and the original question; details are provided in Appendix~\ref{app:evaluation}. \textbf{Indoor EQA} and \textbf{Outdoor EQA} are represented by CoV~\citep{zhao2026cov} and CityEQA~\citep{zhao2025cityeqa}, respectively.

\subsection{Comparison with State of the Art}
\label{subsec:sota}

To answer \textbf{RQ1}, we evaluate end-to-end active perception performance. Results are summarized in Table~\ref{tab:sota}. As shown, Passive QA baselines obtain low QAS and QAC, indicating that language priors or the initial view alone rarely contain sufficient answer evidence. General VLM and policy-only baselines can sometimes localize the target, as reflected by the 57.41\% TDR of Qwen2-VL-2B, but their terminal viewpoints remain insufficient for reliable fine-grained EQA. Existing indoor and outdoor EQA methods improve over generic navigation policies, yet the best non-ScoutVLA baseline reaches only 11.79\% SSR and 16.96\% QAC. In contrast, ScoutVLA-Pi05 achieves 71.90\% SSR, 80.99\% TDR, and 72.70\% QAC, while ScoutVLA-Pi0 obtains the lowest terminal distance. \textbf{\textit{Question-conditioned continuous perception significantly improves both missing evidence acquisition and final EQA success over all existing paradigms.
}}

\begin{table}[t]
    \centering
    \caption{End-to-end active perception performance on FG-EQA. Values report mean $\pm$ standard deviation. Higher is better for SSR, TDR, QAS, and QAC; lower is better for DT. ATL measures path length and should be interpreted together with SSR and TDR.}
    \label{tab:sota}
    \resizebox{\linewidth}{!}{%
    \begin{tabular}{@{}llcccccc@{}}
        \toprule
        Type & Method & SSR (\%) $\uparrow$ & TDR (\%) $\uparrow$ & DT (m) $\downarrow$ & ATL (m) & QAS $\uparrow$ & QAC (\%) $\uparrow$ \\
        \midrule
        Random & Random & 9.86 $\pm$ 0.57 & 47.57 $\pm$ 4.19 & 33.80 $\pm$ 8.53 & 23.92 $\pm$ 12.41 & 1.70 $\pm$ 0.07 & 13.58 $\pm$ 1.18 \\
        \midrule
        Passive QA & Blind (V4 Pro)~\citep{deepseekai2026deepseekv4} & -- & -- & -- & -- & 1.26 $\pm$ 0.11 & 6.60 $\pm$ 1.34 \\
         & Blind (V4 Flash)~\citep{deepseekai2026deepseekv4} & -- & -- & -- & -- & 1.13 $\pm$ 0.08 & 3.30 $\pm$ 0.82 \\
         & Initial (Qwen3-VL)~\citep{bai2025qwen3vl} & -- & -- & -- & -- & 1.50 $\pm$ 0.16 & 9.70 $\pm$ 2.13 \\
        \midrule
        General VLM & Qwen2-VL-2B~\citep{wang2024qwen2vl} & 10.35 $\pm$ 1.87 & \underline{57.41 $\pm$ 3.82} & 40.60 $\pm$ 9.21 & 6.10 $\pm$ 2.19 & 1.60 $\pm$ 0.19 & 13.00 $\pm$ 2.48 \\
         & Qwen3-VL-8B~\citep{bai2025qwen3vl} & 5.73 $\pm$ 1.23 & 37.04 $\pm$ 4.47 & 42.90 $\pm$ 10.42 & 45.30 $\pm$ 14.86 & 1.36 $\pm$ 0.13 & 6.83 $\pm$ 1.46 \\
         & Qwen3-VL~\citep{bai2025qwen3vl} & 2.86 $\pm$ 0.98 & 22.22 $\pm$ 3.11 & 76.35 $\pm$ 16.54 & 102.20 $\pm$ 28.63 & 1.30 $\pm$ 0.14 & 6.39 $\pm$ 1.28 \\
        \midrule
        Navigation & OpenFly~\citep{gao2025openfly} & 4.63 $\pm$ 1.09 & 19.91 $\pm$ 2.87 & 65.50 $\pm$ 14.18 & 72.20 $\pm$ 18.37 & 1.48 $\pm$ 0.17 & 10.57 $\pm$ 2.14 \\
         & TravelUAV~\citep{wang2024traveluav} & 0.00 $\pm$ 0.00 & 12.96 $\pm$ 2.41 & \underline{26.60 $\pm$ 7.83} & 46.00 $\pm$ 12.52 & 1.17 $\pm$ 0.11 & 3.08 $\pm$ 0.77 \\
        \midrule
        Aerial VLA & UAV-Flow~\citep{wang2025uavflow} & 5.73 $\pm$ 1.38 & 29.17 $\pm$ 3.63 & 79.70 $\pm$ 15.82 & 93.40 $\pm$ 24.54 & 1.49 $\pm$ 0.18 & 9.03 $\pm$ 1.81 \\
        \midrule
        Indoor EQA & CoV~\citep{zhao2026cov} & 10.79 $\pm$ 2.16 & 36.11 $\pm$ 4.23 & 52.80 $\pm$ 11.38 & 44.90 $\pm$ 13.24 & \underline{1.80 $\pm$ 0.21} & \underline{16.96 $\pm$ 2.78} \\
        Outdoor EQA & CityEQA~\citep{zhao2025cityeqa} & \underline{11.79 $\pm$ 1.97} & 39.20 $\pm$ 3.92 & 35.70 $\pm$ 8.57 & 10.90 $\pm$ 3.44 & 1.43 $\pm$ 0.14 & 12.11 $\pm$ 2.26 \\
        \midrule
        Ours & ScoutVLA-Pi0 & 60.78 $\pm$ 3.48 & 76.10 $\pm$ 2.89 & \textbf{16.20 $\pm$ 3.16} & 58.95 $\pm$ 7.81 & 3.45 $\pm$ 0.24 & 61.20 $\pm$ 3.58 \\
         & \textbf{ScoutVLA-Pi05} & \textbf{71.90 $\pm$ 2.83} & \textbf{80.99 $\pm$ 2.18} & 17.56 $\pm$ 2.76 & 62.04 $\pm$ 6.54 & \textbf{3.91 $\pm$ 0.19} & \textbf{72.70 $\pm$ 2.92} \\
        \bottomrule
    \end{tabular}}
    \vspace{-1.5em}
    \parbox{\linewidth}{\raggedright\footnotesize ``--'' indicates that the metric is not applicable because the method does not perform navigation. Bold values denote the best overall results, and underlined values denote the best non-ScoutVLA baselines.\par}
\end{table}

\subsection{Ablation on Training Stages and Components}
\label{subsec:ablation}

To address \textbf{RQ2}, we analyze the efficacy of the two-stage recipe and the language-recovery components, as presented in Table~\ref{tab:stage_ablation}. Stage~1 learns strong viewpoint-control policies, but it fundamentally lacks calibrated answer generation. Following Stage~2, ScoutVLA-Pi05 reaches 71.90\% SSR and 72.70\% QAC while retaining an 80.99\% TDR. Removing LoRA adaptation keeps navigation almost unchanged but reduces QAC explicitly to 12.32\%, showing that parameter-efficient semantic reasoning is critical for the secondary QA task. Removing knowledge insulation degrades both TDR and QA, directly validating the action-language interference dilemma when the two divergent objectives are mutually allowed to overwrite parameters. Removing LMAdapter preserves navigation robustly, but abolishes calibrated short-answer generation under the evaluation protocol. \textbf{\textit{
The two-stage decoupled recipe alongside its LMAdapter/LoRA components fundamentally mitigates catastrophic forgetting, seamlessly bridging viewpoint control and answer generation.
}}

\begin{table}[t]
    \centering
    \caption{Ablation on training stages and Stage~2 components.}
    \label{tab:stage_ablation}
    \resizebox{\linewidth}{!}{%
    \begin{tabular}{@{}lcccccc@{}}
        \toprule
        \textbf{Model} & SSR (\%) $\uparrow$ & TDR (\%) $\uparrow$ & DT (m) $\downarrow$ & ATL (m) & QAS $\uparrow$ & QAC (\%) $\uparrow$ \\
        \midrule
        \multicolumn{7}{@{}l}{\textit{Effect of two-stage training}} \\
        ScoutVLA-Pi0-Stage1 & -- & 71.91 $\pm$ 2.95 & 18.06 $\pm$ 3.21 & 60.72 $\pm$ 7.54 & -- & -- \\
        ScoutVLA-Pi0-Stage2 & 61.78 $\pm$ 3.48 & 76.10 $\pm$ 2.89 & \textbf{16.20 $\pm$ 3.16} & 58.95 $\pm$ 7.81 & 3.45 $\pm$ 0.24 & 61.20 $\pm$ 3.58 \\
        \midrule
        ScoutVLA-Pi05-Stage1 & -- & 76.04 $\pm$ 2.51 & 18.01 $\pm$ 2.94 & 62.07 $\pm$ 6.82 & -- & -- \\
        \textbf{ScoutVLA-Pi05-Stage2} & \textbf{71.90 $\pm$ 2.83} & \textbf{80.99 $\pm$ 2.18} & 17.56 $\pm$ 2.76 & 62.04 $\pm$ 6.54 & \textbf{3.91 $\pm$ 0.19} & \textbf{72.70 $\pm$ 2.92} \\
        \midrule
        \multicolumn{7}{@{}l}{\textit{Ablation on Stage~2 components based on ScoutVLA-Pi05}} \\
        w/o LoRA & 10.28 $\pm$ 3.31 & 80.85 $\pm$ 2.15 & 17.61 $\pm$ 2.79 & 62.11 $\pm$ 6.48 & 2.48 $\pm$ 0.26 & 12.32 $\pm$ 1.47 \\
        w/o knowledge insulation & 8.04 $\pm$ 3.52 & 52.12 $\pm$ 3.42 & 22.43 $\pm$ 4.51 & 58.12 $\pm$ 8.17 & 1.48 $\pm$ 0.26 & 10.32 $\pm$ 1.47 \\
        w/o LMAdapter & -- & 80.52 $\pm$ 2.21 & 17.65 $\pm$ 2.84 & 61.98 $\pm$ 6.63 & -- & -- \\
        \bottomrule
    \end{tabular}}
    \vspace{-1.5em}
    \parbox{\linewidth}{\raggedright\footnotesize ``--'' indicates that the model does not provide calibrated short-answer generation, so SSR and QA metrics are not applicable.\par}
\end{table}

\subsection{Terminal-View Answerability and VQA Quality}
\label{subsec:quality_analysis}

\begin{wraptable}{r}{0.56\textwidth}
    \vspace{-1.5em} 
    \centering
    \caption{Terminal-view answerability under a shared VQA panel.}
    \label{tab:terminal_view_quality}
    \resizebox{\linewidth}{!}{%
    \begin{tabular}{@{}llcccc@{}}
        \toprule
        View source & Metric & Qwen3-VL-8B & Qwen3-VL-32B & Qwen3-VL & Mean \\
        \midrule
        \multirow{2}{*}{Qwen2-VL-2B} & QAS & 1.42 & 1.45 & 1.50 & 1.46 \\
         & QAC (\%) & 8.80 & 7.70 & 9.70 & 9.30 \\
        \multirow{2}{*}{Qwen3-VL-8B} & QAS & 1.36 & 1.42 & 1.36 & 1.40 \\
         & QAC (\%) & 6.83 & 7.70 & 6.80 & 7.61 \\
        \multirow{2}{*}{Qwen3-VL} & QAS & 1.11 & 1.17 & 1.30 & 1.22 \\
         & QAC (\%) & 2.20 & 2.00 & 6.39 & 3.96 \\
        \multirow{2}{*}{ScoutVLA-Pi0} & QAS & 2.02 & 2.36 & 2.75 & 2.38 \\
         & QAC (\%) & 25.60 & 33.90 & 43.80 & 34.43 \\
        \multirow{2}{*}{ScoutVLA-Pi05} & QAS & \textbf{2.16} & \textbf{2.56} & \textbf{2.82} & \textbf{2.51} \\
         & QAC (\%) & \textbf{28.90} & \textbf{34.30} & \textbf{45.50} & \textbf{36.27} \\
        \bottomrule
    \end{tabular}}
    \vspace{-1em} 
\end{wraptable}

To answer \textbf{RQ3}, we decouple the evaluation of the generated terminal views from the inherent language-side answering capability. Table~\ref{tab:terminal_view_quality} first evaluates terminal-view answerability. We use Qwen2-VL and Qwen3-VL variants as external VQA answerers. For each view source, we hold the terminal FPV image fixed, ask the same external VQA model panel to answer the original question, and score the generated answers with the standard judge protocol. As shown, ScoutVLA-Pi05 achieves the highest mean QAS and QAC, indicating that its policy produces terminal views with significantly more complete and discriminative visual evidence compared to other baselines.

\begin{wraptable}{r}{0.3\textwidth}
    \vspace{-1.5em}
    \centering
    \caption{Language-side VQA quality.}
    \label{tab:vqa_quality}
    \resizebox{\linewidth}{!}{%
    \begin{tabular}{@{}lcc@{}}
        \toprule
        \textbf{Model} & \textbf{QAS} & \textbf{QAC (\%)} \\
        \midrule
        Qwen2-VL-2B & 3.21 & 50.40 \\
        Qwen3-VL & 4.13 & 77.50 \\
        ScoutVLA-Pi0 & 4.67 & 92.10 \\
        ScoutVLA-Pi05 & \textbf{4.78} & \textbf{95.40} \\
        \bottomrule
    \end{tabular}}
    \vspace{-1em}
\end{wraptable}

Furthermore, Table~\ref{tab:vqa_quality} isolates language-side VQA ability on held-out UAV-scene VQA samples. Qwen2-VL-2B is included as a comparable-scale general VLM baseline, while Qwen3-VL serves as a strong general-purpose reference. ScoutVLA-Pi05 reaches 4.78 QAS and 95.40\% QAC, outperforming both baselines. This indicates that domain-specific training on FG-EQA improves visual-language grounding beyond what is obtained from general pretraining alone. \textbf{\textit{
ScoutVLA's superiority is twofold---it acquires stronger inherent VQA decoding via domain alignment, and more importantly, its active refinement definitively yields evidence-rich terminal perceptions that dramatically raise answerability for any downstream QA model.
}}

\subsection{Real-World Qualitative Deployment}
\label{subsec:real_world}

\begin{figure}[t]
    \centering
    \includegraphics[width=\linewidth]{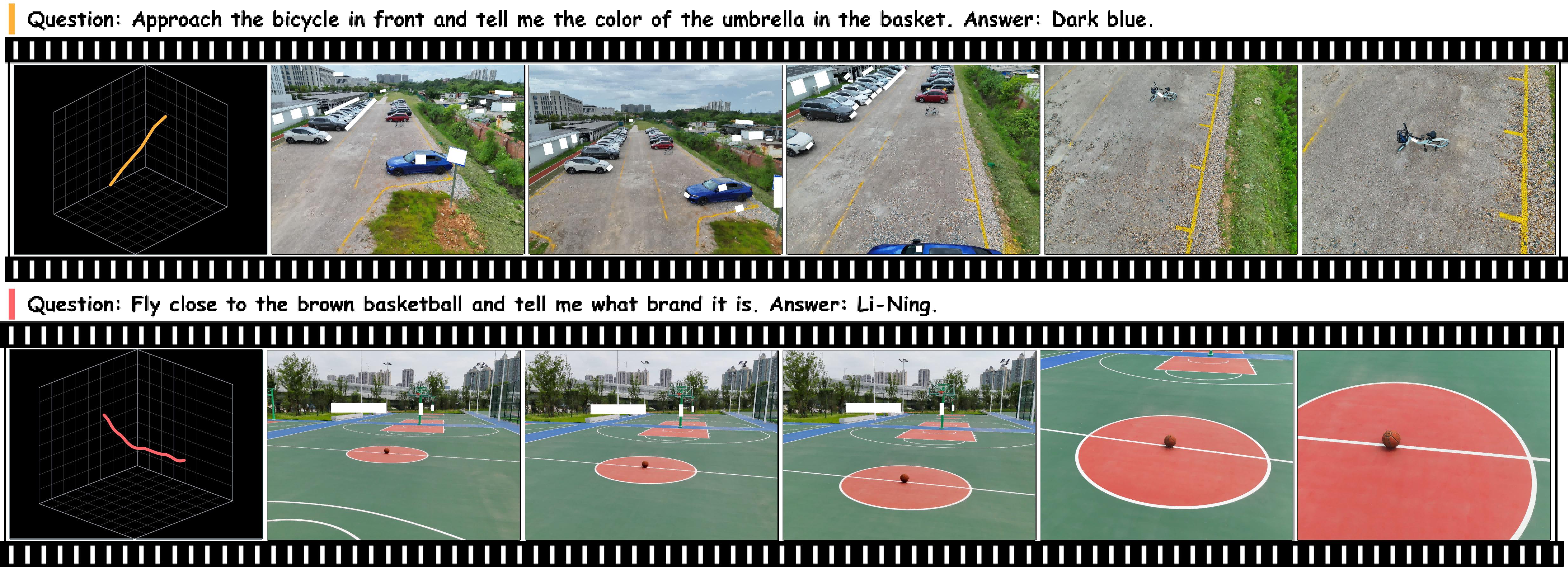}
    \caption{\textbf{Real-world closed-loop deployment.}}
    \label{fig:real_world_deployment}
    \vspace{-1.5em}
\end{figure}

Due to UAV safety constraints and outdoor variability, quantitative evaluation in physical environments remains challenging. Nevertheless, to address \textbf{RQ4} and verify deployability, we fine-tune ScoutVLA-Pi05 on real-world trajectories to bridge the sim-to-real gap. During closed-loop inference, ScoutVLA relies solely on onboard visual streams to execute autonomous continuous flight. As shown in Figure \ref{fig:real_world_deployment}, the policy successfully navigates to the precise overhead close-up, exposing the necessary visual evidence to correctly answer the language query. \textbf{\textit{ScoutVLA effectively generalizes its fine-grained active perception and QA capabilities to real-world hardware.}}

\section{Limitations}
\label{subsec:limitations}

While ScoutVLA significantly advances late-stage active perception, several structural limitations remain. First, the policy assumes the target is already within the general field of view, making it a complement to, rather than a replacement for, macroscopic long-horizon target search pipelines. Second, the current FG-EQA benchmark predominantly features static targets and single-agent trajectories; scaling to highly dynamic outdoor entities, extreme weather conditions, or multi-UAV collaboration requires further investigation. Third, although our qualitative physical deployment shows strong promise, the real-world dataset remains limited, necessitating large-scale physical benchmarking to fully characterize visual sim-to-real robustness. Finally, our 5-DoF continuous action space inherently constrains drone roll to maintain camera stability, whereas more agile 6-DoF maneuvers could be explored in constraint-relaxed flight scenarios.

\section{Conclusion}
\label{sec:conclusion}

In aerial EQA, existing paradigms prematurely halt upon macroscopic target localization, leaving the fine-grained active perception required to gather hidden visual evidence largely unresolved. To address this critical gap, we introduce the FG-EQA benchmark alongside ScoutVLA. Drawing inspiration from the ``waggle dance'' of scout bees, we closely emulate this behavior by coupling a multimodal expert for semantic intent inference with a continuous flow-matching expert for spatial viewpoint generation. To effectively realize this architecture without suffering from action-induced catastrophic forgetting, we employ a decoupled two-stage training paradigm equipped with knowledge insulation. Both rigorous simulation scaling and qualitative real-world deployments explicitly confirm that this design successfully guides UAVs to actively maneuver and expose occluded visual cues. Meaningfully, ScoutVLA shifts the paradigm of aerial EQA from passive geometric observation to proactive, physically-grounded semantic probing. Future work will integrate this active policy with long-horizon search agents and extend the system to support dynamic open-world targets.

\clearpage
\input{appendiex}

\end{document}

%% file: appendiex.tex
\appendix
\section{Appendix Roadmap}
\label{app:roadmap}

This supplementary document provides implementation specifications, analytical frameworks, and qualitative observations corresponding to the main manuscript. The appendix is structured as follows:

\begin{itemize}
    \item \textbf{Appendix~\ref{app:related_vla}} provides an extended discussion of the relevant literature and a detailed architectural taxonomy of recent aerial VLA models.
    \item \textbf{Appendix~\ref{app:dataset}} details the comprehensive construction pipeline of the FG-EQA benchmark, including state-action encodings, rigorous quality-filtering protocols, and descriptive dataset statistics.
    \item \textbf{Appendix~\ref{app:model}} elaborates on the ScoutVLA dual-expert architecture, explicitly detailing the network configurations.
    \item \textbf{Appendix~\ref{app:training}} specifies the exact engineering parameters required to reproduce our two-stage decoupled training paradigm, the gradient insulation mechanism, and the inference workflow.
    \item \textbf{Appendix~\ref{app:evaluation}} defines the formalized evaluation metrics, standardizes the baseline testing procedures, and provides the exact prompt templates utilized by our LLM judge panel.
    \item \textbf{Appendix~\ref{app:experiments_deployment}} details the real-world hardware deployment architecture, provides supplementary experimental analyses and discusses qualitative success cases alongside recognized failure modes in both simulation and physical deployments.

\end{itemize}

\vspace{1em}
\noindent\textbf{Anonymous Resources}

To facilitate the peer-review process and ensure reproducibility, we have released our anonymized project repository and a dedicated project website. The repository includes the core source code for ScoutVLA, the main manuscript, and the supplementary appendix. The project website hosts qualitative demonstration videos. The full FG-EQA benchmark dataset (including all simulated and real-world trajectories) will be open-sourced and made publicly available in the near future. 

The anonymized resources can be temporarily accessed at:
\begin{itemize}\raggedright
    \item \textbf{Anonymous Repository:} \url{https://anonymous.4open.science/r/ScoutVLA-9657}
    \item \textbf{Anonymous Project Website:} \url{https://anonymous.4open.science/w/ScoutVLA-B31F/}
\end{itemize}

\section{Extended Related Work Notes}
\label{app:related_vla}

This section provides an extended discussion of the related literature and a detailed architectural comparison of recent aerial VLA models, which is abbreviated in the main text due to space constraints.

\textbf{Embodied question answering.}
Early EQA studies focus mainly on indoor environments, where navigation and question answering are tightly coupled~\citep{das2017embodiedqa,majumdar2024openeqa,cheng2025efficienteqa,zhai2025memory,jiang2025beyond}. The transition to outdoor EQA introduced vastly larger spatial scales and sparser targets, leading recent systems to adopt modular pipelines. Pioneering works like CityEQA~\citep{zhao2025cityeqa}, BridgeEQA~\citep{varghese2025bridgeeqa}, and IndustryEQA~\citep{li2025industryeqa} have made significant strides in macroscopic target search and high-level spatial reasoning. However, these works implicitly assume that reaching the target's general vicinity equates to collecting sufficient visual evidence for the query. Due to the lack of the critical fine-grained viewpoint adjustment process, these agents often generate incorrect answers when confronted with questions requiring specific, angle-dependent observations. ScoutVLA specifically targets this overlooked active perception phase prior to final QA.

\textbf{Active perception.}
The core philosophy of active perception is that an agent can maximize information gain by autonomously altering its sensor state. Recently, this concept has been explored primarily from the perspectives of macroscopic geometric exploration and model cognitive evolution. For instance, LongNav-R1~\citep{hu2026longnav} conceptualizes navigation as a multi-turn interaction between policy and environment, utilizing reinforcement learning for adaptive exploration. AirCopBench~\citep{zha2025aircopbench} establishes a comprehensive benchmark for multi-UAV collaborative perception under degraded conditions. Meanwhile, Active-Zero~\citep{he2026activezero} enables active self-evolution of Vision-Language Models through a multi-agent collaborative framework. However, a recent systematic evaluation of spatial reasoning reveals a fundamental bottleneck in current methods: severe spatial knowledge degradation caused by belief inertia and inefficient exploration. This indicates that existing work largely remains at the level of macroscopic geometric exploration, lacking the capability for micro-viewpoint mapping driven by complex semantics. In contrast, ScoutVLA advances active perception from macroscopic target search to question-conditioned fine-grained view finding, utilizing continuous 5-DoF control to fulfill precise observations within minimal visual tolerances.

\textbf{Aerial vision-language-action models.}
By integrating the reasoning capabilities of Large Language Models with low-level motion control, Vision-Language-Action (VLA) models have gained significant traction and are increasingly applied to the UAV domain. Systems such as OpenFly~\citep{gao2025openfly}, FlightGPT~\citep{cai2025flightgpt}, and TravelUAV~\citep{wang2024traveluav} are primarily engineered as macroscopic navigation models, focusing on language-guided long-range waypoint planning rather than target-focused observation. Conversely, short-range control models like SkyVLN~\citep{li2025skyvln}, UAV-Flow~\citep{wang2025uavflow}, and VLA-AN~\citep{wu2025vlaan} have validated the feasibility of diffusion-style or continuous action heads for flight, yet they remain dedicated to path following rather than semantics-driven viewpoint mapping. Despite utilizing active mobility, these aerial VLA agents fundamentally lack the capacity for micro-viewpoint refinement driven by complex queries. 

To illustrate this architectural distinction, Table~\ref{tab:vla_models} summarizes the landscape of recent aerial VLA models. Existing baselines are primarily designed for destination arrival. Consequently, they heavily rely on fixed camera configurations, process only simple navigational commands, and yield discrete or lower-DoF actions without deep integration of question answering and flight control. Unlike existing methods, ScoutVLA exclusively uses a monocular RGB camera and integrates the linguistic query directly into the action-generation context through a dual-expert architecture and flow matching paradigm, enabling continuous 5-DoF fine-grained control to actively reveal occluded visual evidence.

\begin{table}[t]
    \centering
    \caption{Architectural comparison of recent VLA models for UAVs. Compared to existing baselines, ScoutVLA is uniquely designed for continuous fine-grained active perception driven by complex semantic questions, while replacing the traditional autoregressive scheme with flow matching.}
    \label{tab:vla_models}
    \resizebox{\linewidth}{!}{%
    \begin{tabular}{lcccccccccc}
        \toprule
        \textbf{Model} & \textbf{DoF} & \textbf{Trajectory Length (m)} & \textbf{Camera Count} & \textbf{Fixed Camera} & \textbf{Image Type} & \textbf{Action Space} & \textbf{Question Input} & \textbf{Viewpoint} & \textbf{Model Size} & \textbf{Action Head Type} \\
        \midrule
        OpenFly~\citep{gao2025openfly} & 4 & 99 & 1 & \cmark & RGB & Discrete & \xmark & FPV & 7B & Autoregressive \\
        FlightGPT~\citep{cai2025flightgpt} & 4 & 200 & 1 & \cmark & RGB & Discrete & \xmark & Top-Down & 7B & Autoregressive \\
        TravelUAV~\citep{wang2024traveluav} & 6 & 255 & 5 & \cmark & RGB-D & Continuous & \xmark & Multi-view & 7B & Autoregressive \\
        OpenVLA-UAV~\citep{wang2025uavflow} & 6 & 10 & 1 & \cmark & RGB & Discrete & \xmark & FPV & 7B & Autoregressive \\
        Pi0-UAV~\citep{wang2025uavflow} & 6 & 10 & 1 & \cmark & RGB & Continuous & \xmark & FPV & 4B & Flow Matching \\
        VLA-AN~\citep{wu2025vlaan} & 4 & 10 & 1 & \cmark & RGB-D & Continuous & \xmark & FPV & 2B/3B/7B & Autoregressive \\
        \midrule
        \textbf{ScoutVLA (Ours)} & \textbf{5} & \textbf{30} & \textbf{1} & \textbf{\xmark} & \textbf{RGB} & \textbf{Continuous} & \textbf{\cmark} & \textbf{FPV} & \textbf{2B} & \textbf{Flow Matching} \\
        \bottomrule
    \end{tabular}}
\end{table}

\section{FG-EQA Construction Details}
\label{app:dataset}

\subsection{State and Action Encoding}
\label{app:dataset_encoding}

The main paper defines the 5-DoF physical command inline. For implementation, the absolute UAV state is represented as Equation~\ref{eq:state}, where $(x_t,y_t,z_t)$ denotes the world-frame position, $\psi_t$ is the body yaw, and $\theta^g_t$ is the gimbal pitch. We encode the yaw angle utilizing its sine and cosine components to eliminate angular wrap-around discontinuities.

\begin{equation}
    \mathbf{s}_t = [x_t, y_t, z_t, \sin(\psi_t), \cos(\psi_t), \theta^g_t]
    \label{eq:state}
\end{equation}

For model prediction, the action chunk is formulated as relative increments, as shown in Equation~\ref{eq:action}, where the positional increments $(\Delta x_t, \Delta y_t, \Delta z_t)$ are executed relative to the current UAV body frame. $\sigma^{\mathrm{stop}}_t \in [0, 1]$ is a specialized stop score introduced to signal sequence termination. During inference, the UAV halts its active exploration when this score exceeds a threshold of 0.85. This value was established through an extensive empirical search during a validation phase. In our experiments, lower thresholds (e.g., $< 0.8$) often led to premature halting before sufficient visual evidence was acquired. Conversely, higher thresholds (e.g., $> 0.9$) typically triggered over-exploration, causing the UAV to fly excessively close to the target, which increased the risk of losing the target from the field of view or potential collisions. Consequently, the threshold of 0.85 was selected as it achieves the optimal trade-off between spatial navigation efficiency and visual evidence sufficiency.

\begin{equation}
    \tilde{\mathbf{a}}_t =
    [\Delta x_t, \Delta y_t, \Delta z_t, \sin(\Delta \psi_t),
    \cos(\Delta \psi_t), \Delta \theta^g_t, \sigma^{\mathrm{stop}}_t]
    \label{eq:action}
\end{equation}

During inference, both architectural variants output the encoded vector $\tilde{\mathbf{a}}_t$ through flow-matching generation. This internal prediction is subsequently decoded—by recovering the scalar yaw angle via $\operatorname{atan2}(\sin\Delta\psi_t, \cos\Delta\psi_t)$ and separating the stop mechanism—back into the explicit 5-DoF physical command $\mathbf{u}_t$ (defined in Section~\ref{subsec:task}) to be executed by the flight controller.

\subsection{Perception Trajectory Generation}
\label{app:dataset_trajectory}

\begin{figure}[t]
    \centering
    \includegraphics[width=0.9\linewidth]{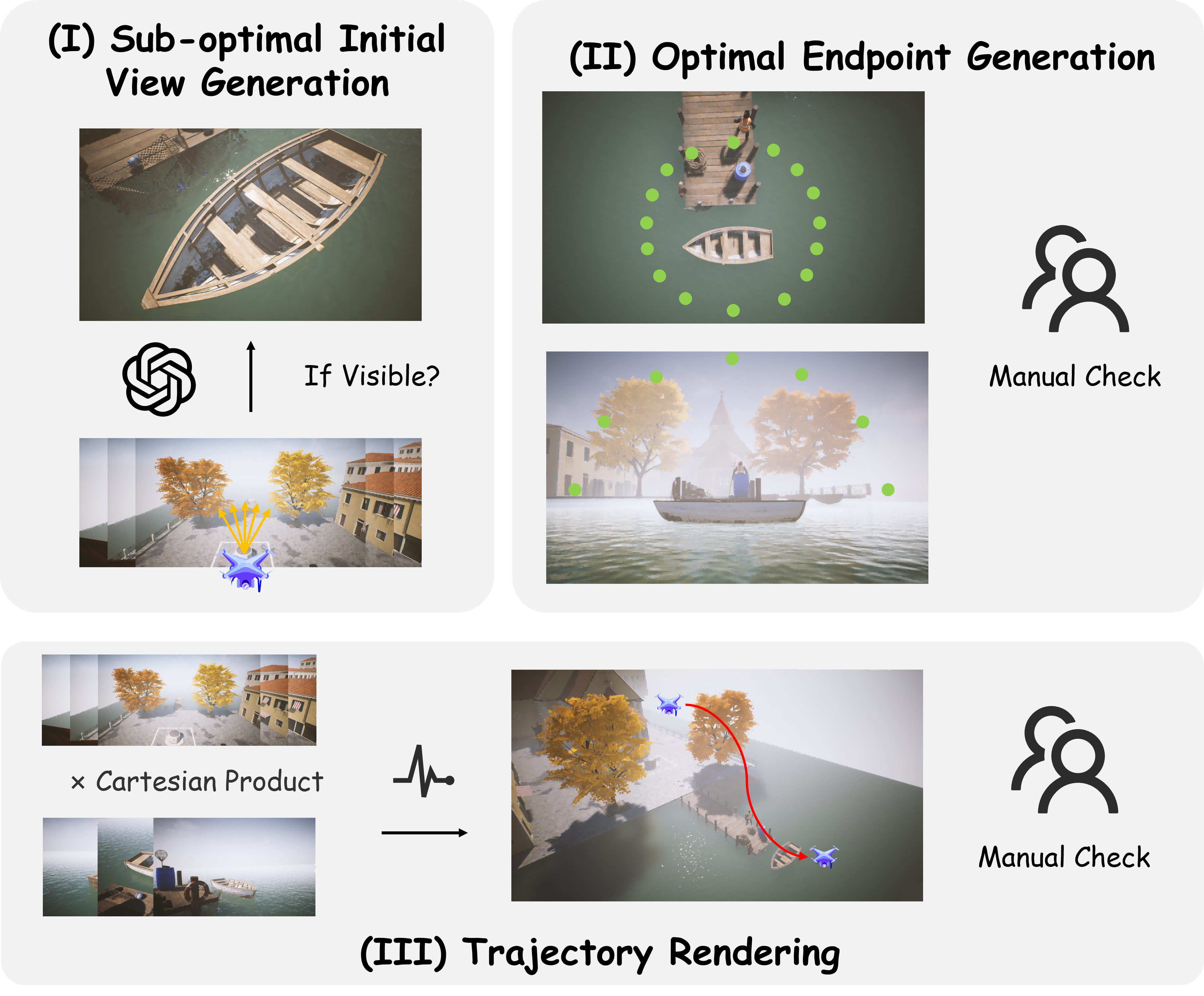}
    \caption{\textbf{The simulated perception trajectory generation pipeline.}}
    \label{fig:dataset_pipeline}
\end{figure}

As introduced in Section~\ref{subsec:data_construction}, the perception trajectory generation phase is designed to formulate the precise viewpoint-refinement flight paths. To systematically scale this process in simulation, we construct these observation trajectories through multi interconnected computational phases (illustrated in Figure~\ref{fig:dataset_pipeline}).

\begin{figure}[t]
    \centering
    \includegraphics[width=0.9\linewidth]{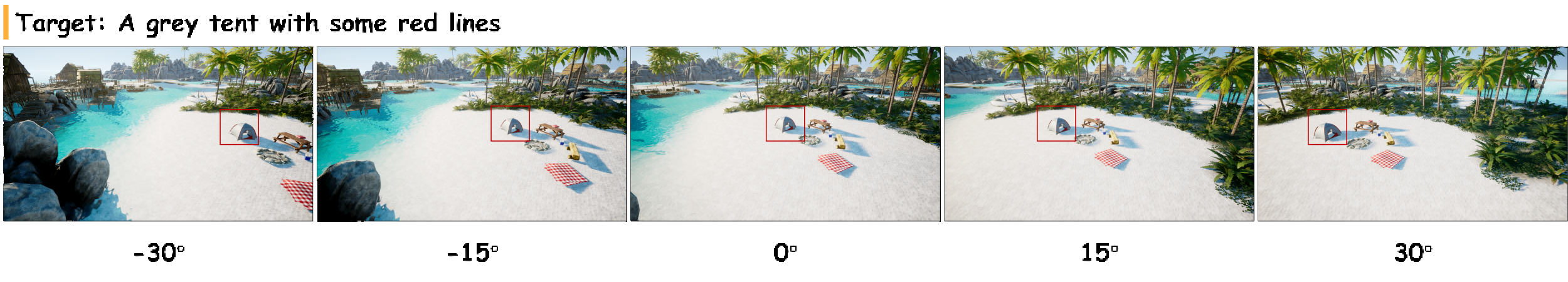}
    \caption{\textbf{Simulated initial pose variations.} The UAV systematically applies yaw offsets of $0^\circ, \pm 15^\circ$, and $\pm 30^\circ$ relative to the line-of-sight, generating 5 distinct directional poses for robustness.}
    \label{fig:initial_5_poses}
\end{figure}

\begin{figure}[t]
    \centering
    \includegraphics[width=0.9\linewidth]{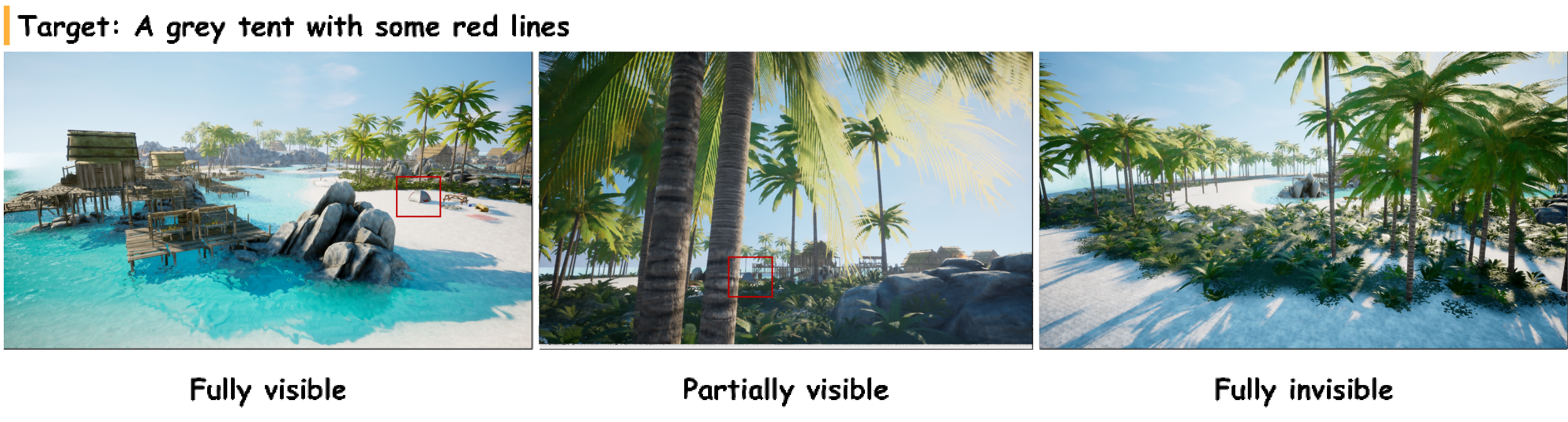}
    \caption{\textbf{Initial view visibility filtering.} Qwen3-VL acts as an automated visual filter, categorizing candidate poses into fully visible, partially visible, and completely invisible. Only the first two categories are retained to simulate realistic macroscopic search termination.}
    \label{fig:initial_filtering}
\end{figure}

\textbf{Phase 1: Sub-optimal Initial View Generation.} To simulate the termination of a macroscopic navigation phase where the target has just entered the Field of View (FOV), we establish the base orientation of the UAV as the direct line-of-sight connecting the starting position to the target. To enrich the dataset and emulate realistic pose variations, we systematically apply yaw offsets of $0^\circ, \pm 15^\circ$, and $\pm 30^\circ$ relative to this base direction, thereby generating 5 distinct directional poses for each starting location (as visually demonstrated in Figure~\ref{fig:initial_5_poses}). Furthermore, due to complex outdoor environments, structural objects between the initial point and the target often cause partial or total occlusions, rendering the target indiscernible. To systematically address this, we employ Qwen3-VL as an automated visual filter to evaluate all candidate start poses. The VLM categorizes each initial view into one of three visibility states: \textit{fully visible}, \textit{partially visible}, and \textit{fully invisible}. We retain only the fully and partially visible instances, discarding the completely invisible ones (examples of this filtering categorization are provided in Figure~\ref{fig:initial_filtering}). This meticulous setup functionally mimics the realistic scenario where a macroscopic search agent hands over control to our fine-grained active perception module under suboptimal viewing angles.

\begin{figure}[t]
    \centering
    \includegraphics[width=0.9\linewidth]{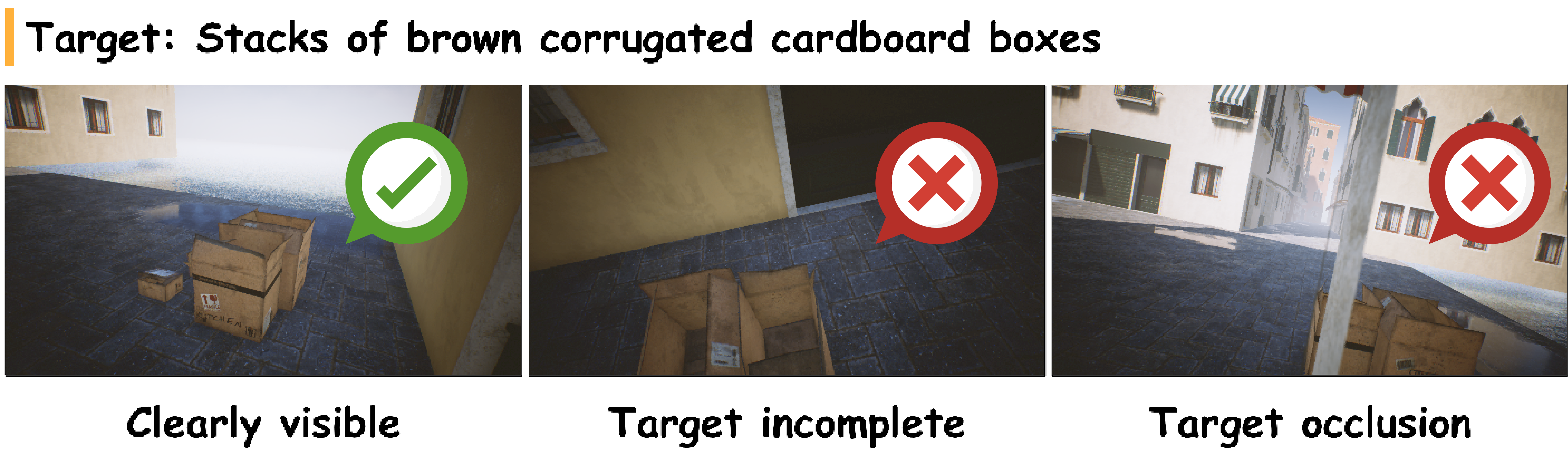}
    \caption{\textbf{Optimal endpoint sampling and expert screening.} Candidate terminal viewpoints are sampled from the upper hemisphere and systematically evaluated by domain experts. Only completely unoccluded and visually clear endpoints are retained.}
    \label{fig:endpoint_filtering}
\end{figure}

\textbf{Phase 2: Optimal Endpoint Generation.} To accurately determine the endpoints and attitudes for the fine-grained UAV trajectories, we generate observation spheres centered on each specific target. The observation radius for each sphere is dynamically adjusted according to the varying physical dimensions of the corresponding target. Candidate observation points are then discretely sampled exclusively across the upper hemisphere. During sampling, we strictly constrain the camera orientation at each point to align precisely with the direct vector connecting the observation coordinate to the target center. While this geometric formulation yielded a diverse set of candidate terminal views, the generated images exhibit varying framing qualities, with some suffering from partial occlusions or incomplete visibility due to complex environmental layouts. To ensure data reliability, we conduct a manual screening process involving a panel of ten researchers with expertise in UAV operations. This expert group evaluate every rendered candidate image. For each target, they carefully filter the pool, retaining only those terminal observation points that provided an entirely clear, unoccluded, and complete view of the target (examples of the candidate endpoint sampling and the expert retention criteria are illustrated in Figure~\ref{fig:endpoint_filtering}).

\textbf{Phase 3: Trajectory Rendering.} Base trajectories are constructed by navigating from the valid sub-optimal initial positions to the sampled endpoints. The flight involves kinematic interpolation, keeping the camera smoothly tracking the target at a 10 Hz recording rate to ensure fluid spatial transition. Crucially, a strict collision detection mechanism is implicitly integrated during this continuous movement. If the UAV collides with any environmental geometry along the interpolated path, that specific trajectory is immediately discarded to ensure absolute physical feasibility.

\textbf{Phase 4: Recovery Perturbation.} To endow the VLA model with robust error-correction capabilities, we inject random body-attitude and spatial perturbations into approximately 30\% of the simulated trajectories. These serve as critical recovery samples during Stage 1 flow-matching training, teaching the policy how to re-center the target if it drifts.

\textbf{Phase 5: Final Manual Verification.} After all automated computational phases, we conduct a concluding manual review across the generated dataset. Human annotators carefully replay and inspect the trajectory recordings to identify and eliminate any remaining defective episodes. Specifically, they screen for subtle unrecorded collisions, degraded visual clarity at either the starting or terminal viewpoints, and any other unpredictable simulation anomalies. This exhaustive human-in-the-loop verification ensures the quality, safety, and visual reliability of the finalized trajectories.

\subsection{QA Data Annotation and Generic VQA Integration}
\label{app:dataset_qa}

As outlined in Section~\ref{subsec:data_construction}, the terminal viewpoints selected by UAV experts in Phase 2 provide definitive visual evidence. These high-quality images act as the explicit visual basis for QA annotation. To systematically generate and refine the QA pairs, we implement a comprehensive pipeline comprising automated generation and a rigorous three-stage filtering mechanism.

\begin{figure}[t]
    \centering
    \includegraphics[width=0.9\linewidth]{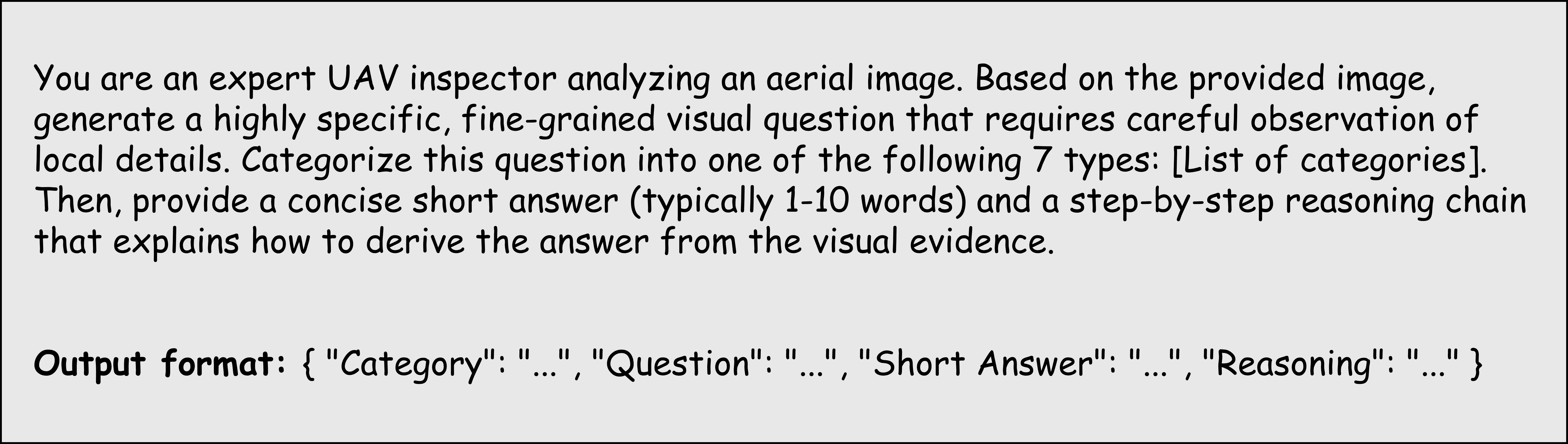}
    \caption{\textbf{System Prompt for QA Generation.} The prompt template designed to instruct the VLM to generate highly specific, fine-grained visual queries and their corresponding reasoning chains based on localized visual evidence.}
    \label{fig:c3_prompt}
\end{figure}

\textbf{Question-Answer Generation.} 
We utilize Qwen3-VL to automatically generate candidate question-answer pairs and their underlying reasoning chains. Taking the optimal terminal FPV image as input, the VLM is instructed to produce queries spanning seven fine-grained evidence categories: \textit{Object Counting}, \textit{State Recognition}, \textit{Defect Detection}, \textit{Spatial Reasoning}, \textit{Attribute Recognition}, \textit{Activity Recognition}, and \textit{Text Recognition}. The prompt is designed to enforce highly localized visual grounding and concise answers, as shown in Figure~\ref{fig:c3_prompt}.

\textbf{Three-Stage Quality Filtering.}
To ensure the generated data strictly assesses active perception rather than mere spatial navigation or language priors, we apply a stringent three-stage filtering protocol to the generated candidates:
\begin{itemize}
    \item \textbf{Stage 1: Multimodal Self-Consistency Verification.} First, to ensure factual correctness, the generated triplet (Question, Reasoning, Answer) is evaluated by Qwen3-VL against the optimal terminal image in an independent session. The model is tasked to rigorously verify whether the question is unambiguously resolvable from the image and whether the answer logically derives from the reasoning chain. Any candidate exhibiting visual hallucination, logical disconnects, or factual ambiguity is immediately discarded.
    \item \textbf{Stage 2: Counterfactual Initial-View Answerability Check.} This is the most critical step to enforce the necessity of fine-grained maneuvering. We present the verified questions to the VLM, but this time paired exclusively with the sub-optimal \emph{initial} view from Phase 1. If the VLM can successfully deduce the correct answer using only this macroscopic, unrefined starting viewpoint, the sample is classified as trivial and discarded. This counterfactual check guarantees that executing continuous viewpoint refinement is necessary to expose the occluded visual evidence.
    \item \textbf{Stage 3: Human-in-the-Loop Validation Fallback.} While LLM-based filtering is highly scalable, it occasionally misses subtle visual-semantic mismatches. As a final safeguard, professional human annotators conduct a manual review of the surviving samples. They specifically eliminate edge cases such as awkward natural language phrasings, overly subjective questions, or scenarios where severe environment occlusions were mistakenly validated by the VLM. This comprehensive human verification ensures QA reliability.
\end{itemize}

\begin{figure}[t]
    \centering
    \includegraphics[width=\linewidth]{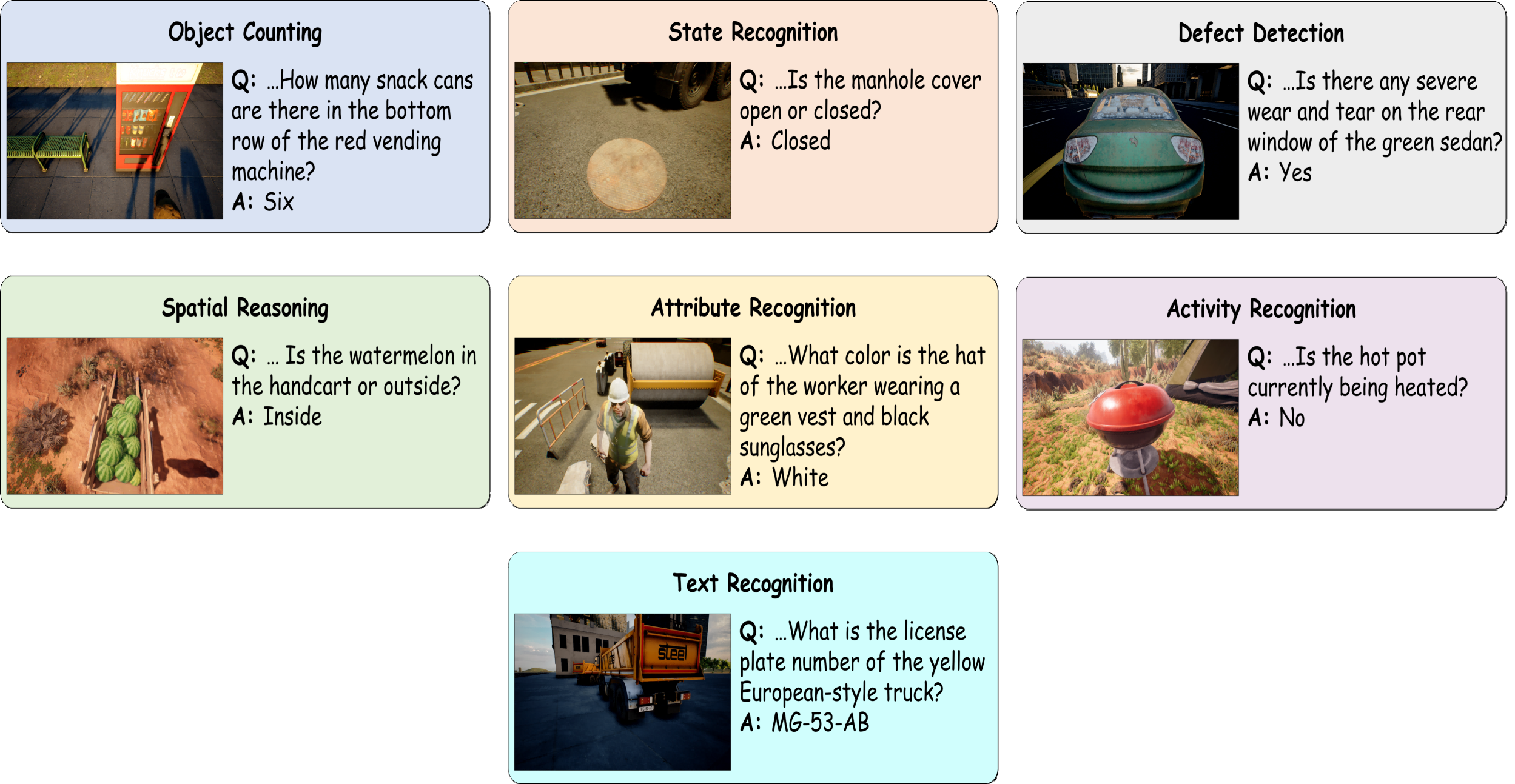}
    \caption{\textbf{Visual examples of the seven fine-grained evidence categories in FG-EQA.}}
    \label{fig:qa_examples}
\end{figure}

Figure~\ref{fig:qa_examples} visually illustrates representative examples from the finalized FG-EQA benchmark across all seven categories, demonstrating the necessity of viewpoint adjustment to acquire the hidden evidence.

\textbf{Generic VQA Integration.} 
Relying exclusively on simulated EQA samples risks severe semantic overfitting to synthetic textures and monotonic 3D assets, which would ultimately cripple the sim-to-real transferability during physical deployment. To mitigate this domain gap and enrich the model's open-world visual vocabulary, we systematically integrate generic real-world VQA samples filtered from the LLaVA-Instruct-150K dataset~\citep{liu2023llava}. However, naively mixing standard conversational data can catastrophically degrade the continuous control capabilities. Therefore, we subject the external dataset to a rigorous three-step semantic-alignment pipeline: (1) \emph{Task Filtering}: exclusively isolating data explicitly categorized as objective visual question answering; (2) \emph{Format Alignment}: excising all multi-turn dialogue structures to strictly enforce a single-turn `Query-to-Evidence' mapping paradigm; and (3) \emph{Length Constraint}: discarding any samples with target responses exceeding a stringent 32-character limit. This final constraint is paramount, as it prevents the model from degenerating into a conversational chatbot, aligning the textual generation head with the concise, factual short-answer logic required for an aerial inspector. This process yields approximately 80K highly refined, domain-generalized instances, which serve as a robust semantic anchor during Stage 2 knowledge-insulated training.

\subsection{Simulation Environments and Real-World Data Collection}
\label{app:dataset_environments}

Before detailing the statistical properties of FG-EQA, we explicitly define the environmental configurations and the physical data collection protocols that undergird our benchmark.

\begin{figure}[t]
    \centering
    \includegraphics[width=\linewidth]{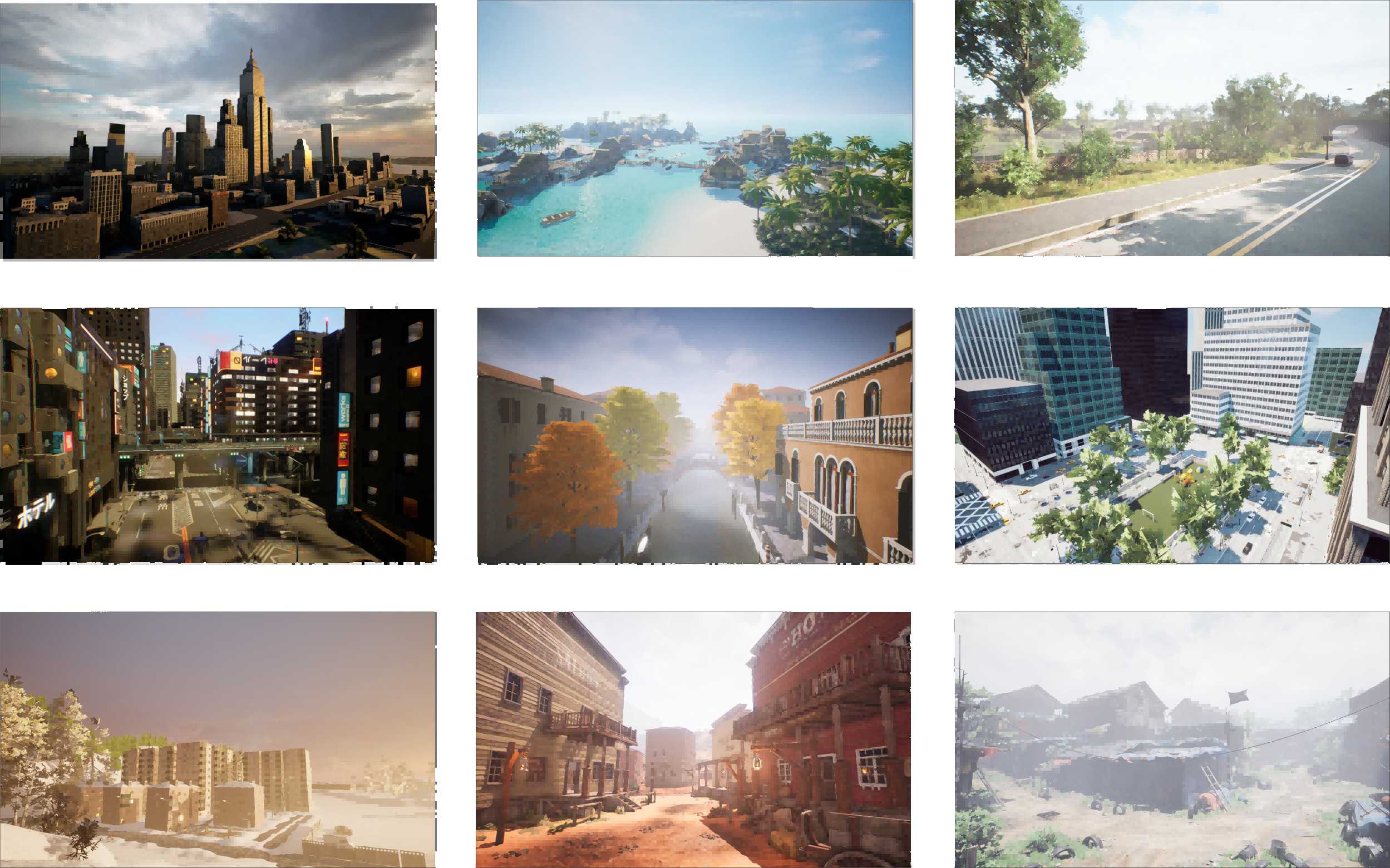}
    \caption{\textbf{The 9 high-fidelity digital-twin simulation environments in FG-EQA.} These environments simulate diverse weather conditions, complex lighting, and dense structural layouts to robustly challenge the agent's fine-grained perception capability.}
    \label{fig:sim_envs}
\end{figure}

\textbf{High-Fidelity Simulation Environments.} To ensure robust multi-scenario generalization, all automated simulated data collection (comprising $\sim$40K episodes) is conducted across 9 high-fidelity digital-twin environments built upon the UAV-ON~\citep{xiao2025uavon} framework, as illustrated in Figure~\ref{fig:sim_envs}. These environments encompass a wide spectrum of semantic layouts, including dense urban blocks, intricate industrial zones, active construction sites, and cluttered residential communities (e.g., \textit{BrushifyUrban}, \textit{CabinLake}, \textit{Slum}). More importantly, these environments are engineered to introduce severe physical and visual constraints. They feature complex structural occlusions (e.g., targets obscured by chain-link fences or tree canopies), dynamic shadow variations driven by shifting sun elevations, and densely packed adversarial distractors. These environmental complexities are deliberately leveraged to obscure initial target observations, thereby strictly necessitating the continuous, fine-grained active perception maneuvers that ScoutVLA must learn to execute.

\begin{figure}[t]
    \centering
    \includegraphics[width=\linewidth]{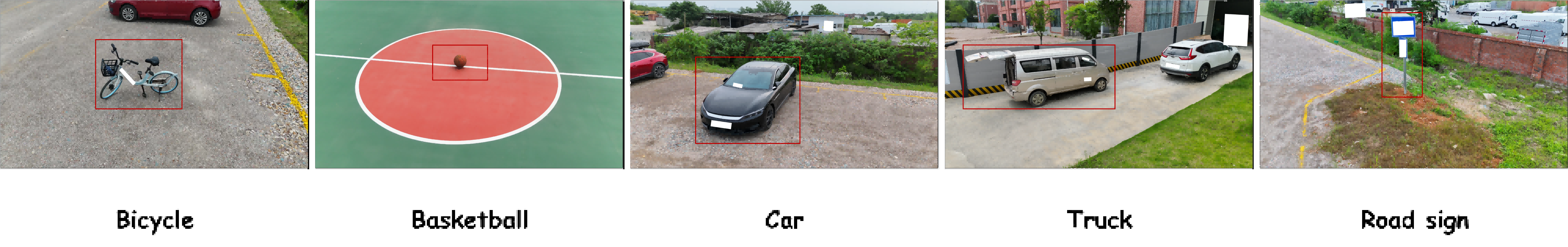} 
    \caption{\textbf{Representative real-world inspection targets.} The 1,000 real-world trajectories cover five distinct categories, capturing authentic sensor noise and unconstrained lighting to validate the real-world applicability of the proposed ScoutVLA architecture.}
    \label{fig:real_envs}
\end{figure}

\textbf{Real-World Data for Dedicated Physical Validation.} Distinct from the massive simulation dataset utilized to train the simulated agent, we explicitly curated a separate real-world split to train and evaluate a dedicated physical-scene policy. This subset comprises 1,000 real-world trajectories, uniformly distributed across five representative physical inspection targets (Bicycle, Basketball, Car, Truck, and Road Sign), with 200 trajectories per task, as illustrated in Figure~\ref{fig:real_envs}. During the data collection phase, professional UAV pilots manually teleoperated the drone, executing micro-viewpoint modifications to continuously gather the hidden visual evidence. These expert demonstrations capture a multitude of authentic physical artifacts absent in simulation, including hardware aerodynamic instability (e.g., wind-induced jitter), unconstrained outdoor lighting shifts, motion blur, and realistic RGB sensor noise (e.g., dynamic range clipping and lens flare). By training a dedicated model strictly on this physical dataset, we empirically demonstrate that the ScoutVLA architectural design natively accommodates real-world complexities, definitively validating its practical applicability beyond synthetic environments.

\subsection{Descriptive Data Analysis}
\label{app:dataset_analysis}

\begin{figure}[t]
    \centering
    \begin{subfigure}[b]{0.24\linewidth}
        \centering
        \includegraphics[width=\linewidth]{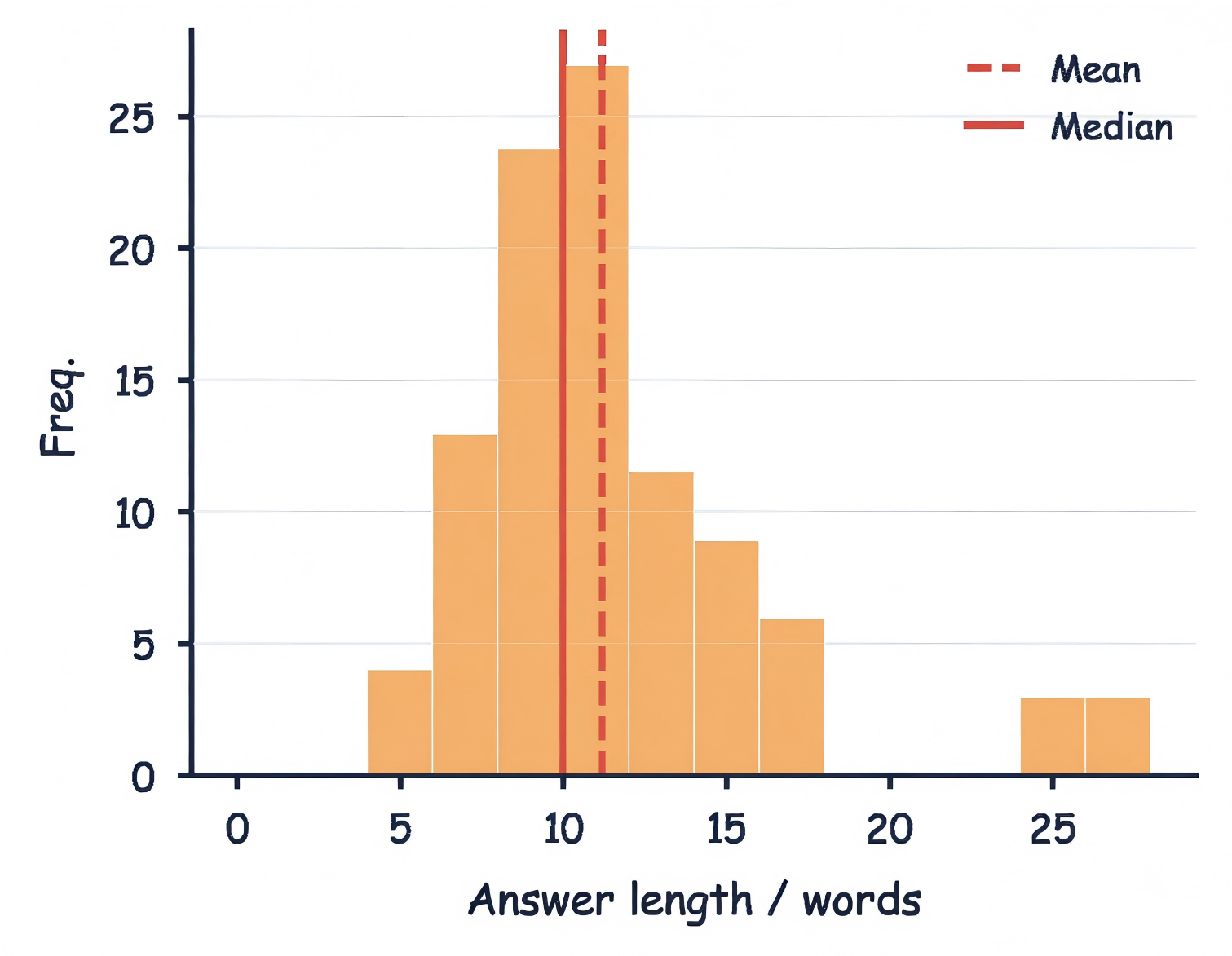}
        \caption{Simulated VQA Len.}
    \end{subfigure}
    \hfill
    \begin{subfigure}[b]{0.24\linewidth}
        \centering
        \includegraphics[width=\linewidth]{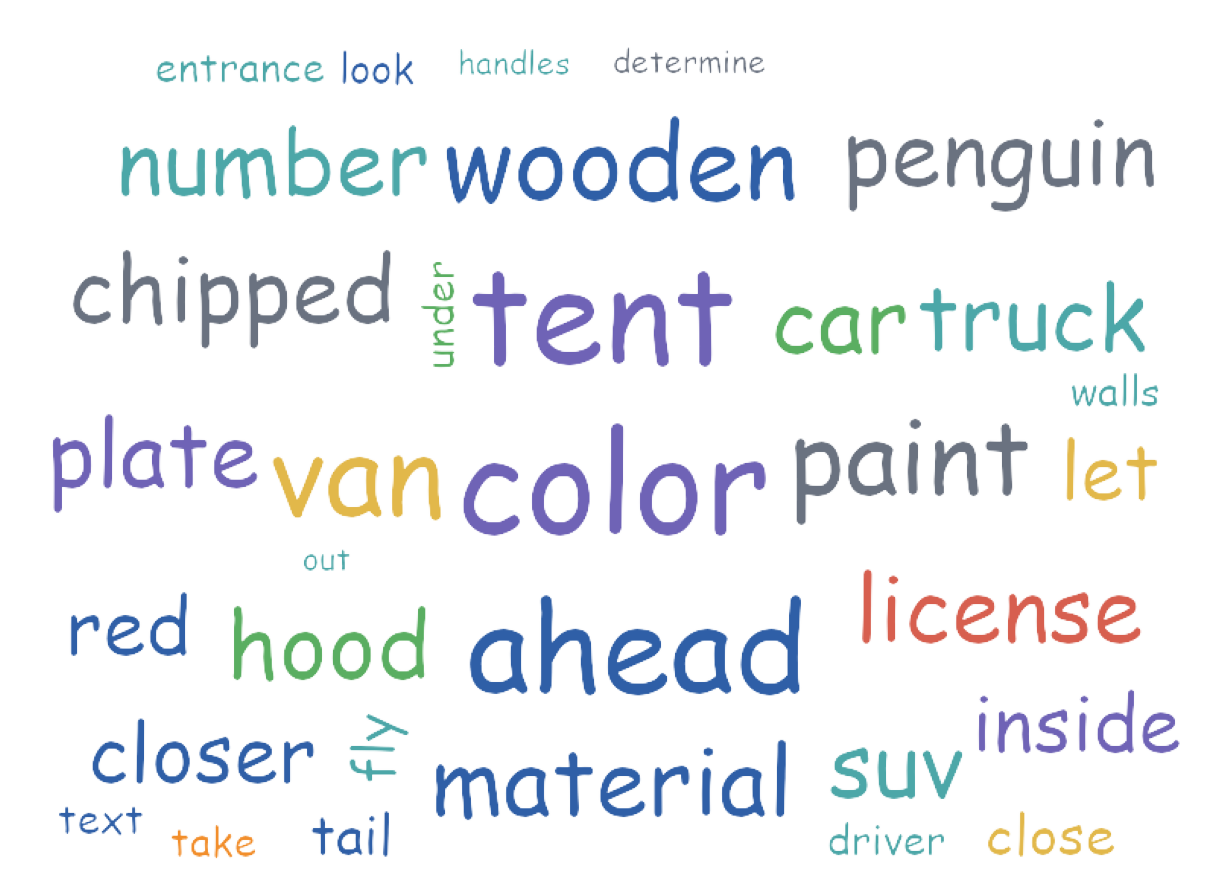}
        \caption{Simulated Word Cloud}
    \end{subfigure}
    \hfill
    \begin{subfigure}[b]{0.24\linewidth}
        \centering
        \includegraphics[width=\linewidth]{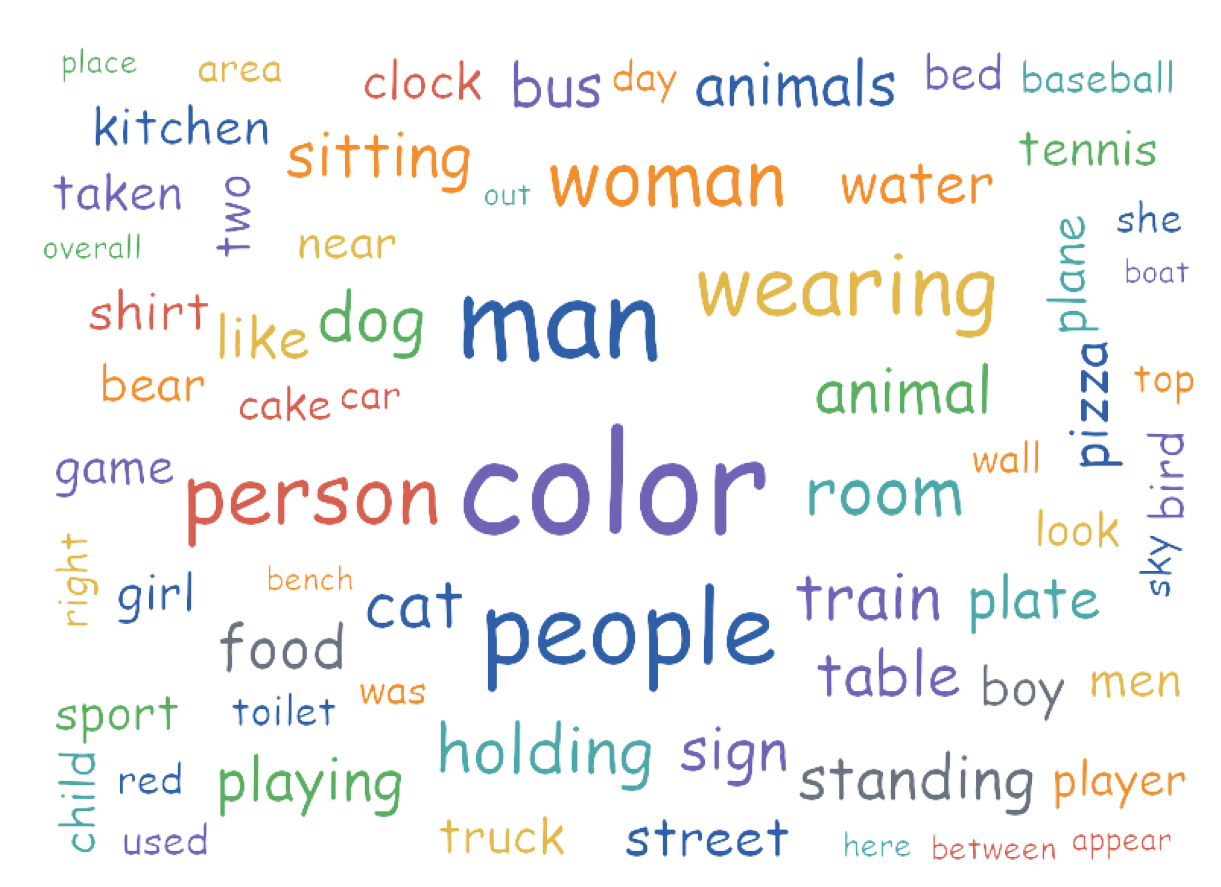}
        \caption{Generic Word Cloud}
    \end{subfigure}
    \hfill
    \begin{subfigure}[b]{0.24\linewidth}
        \centering
        \includegraphics[width=\linewidth]{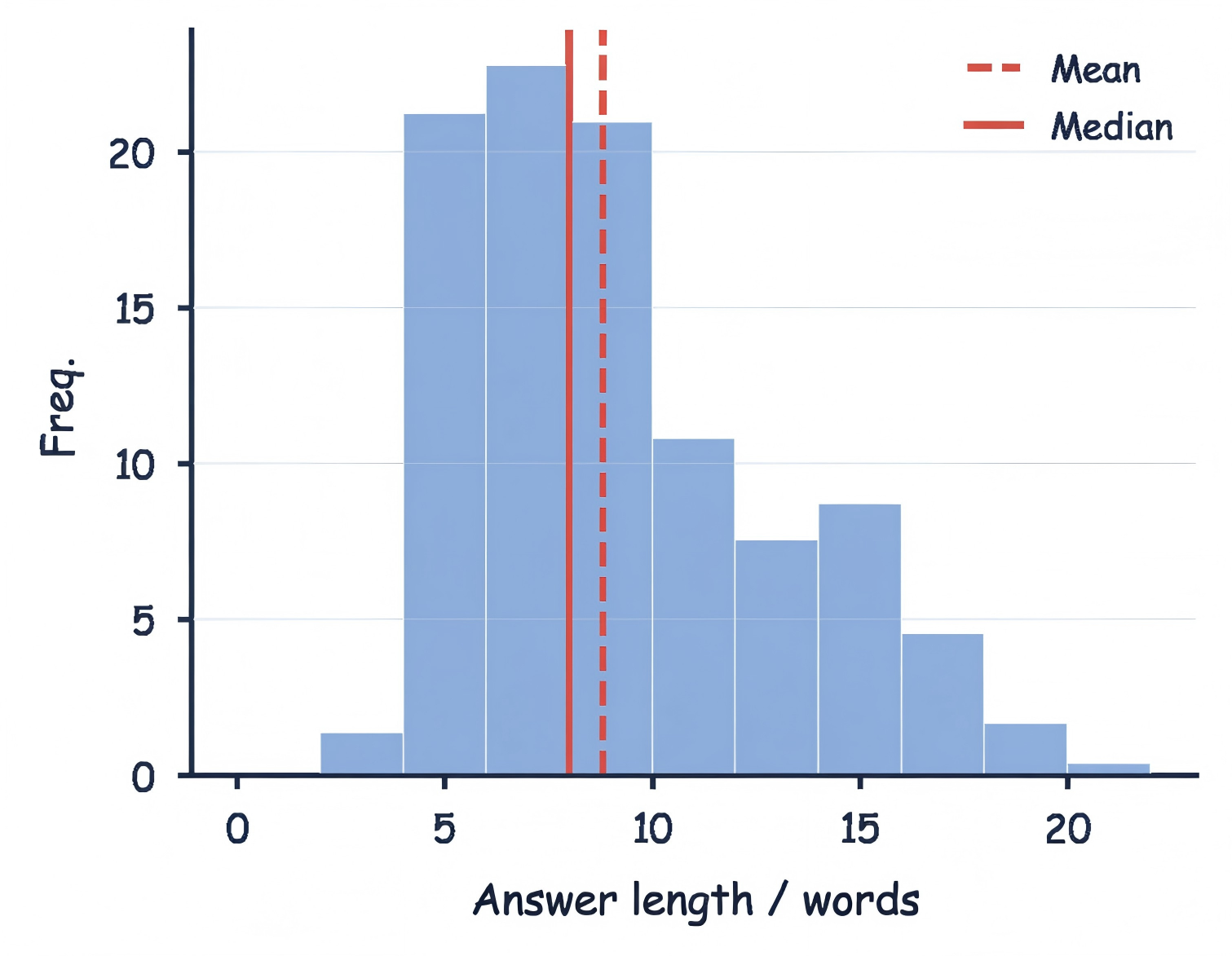}
        \caption{Generic VQA Len.}
    \end{subfigure}

    \vspace{1em} 

    \begin{subfigure}[b]{0.24\linewidth}
        \centering
        \includegraphics[width=\linewidth]{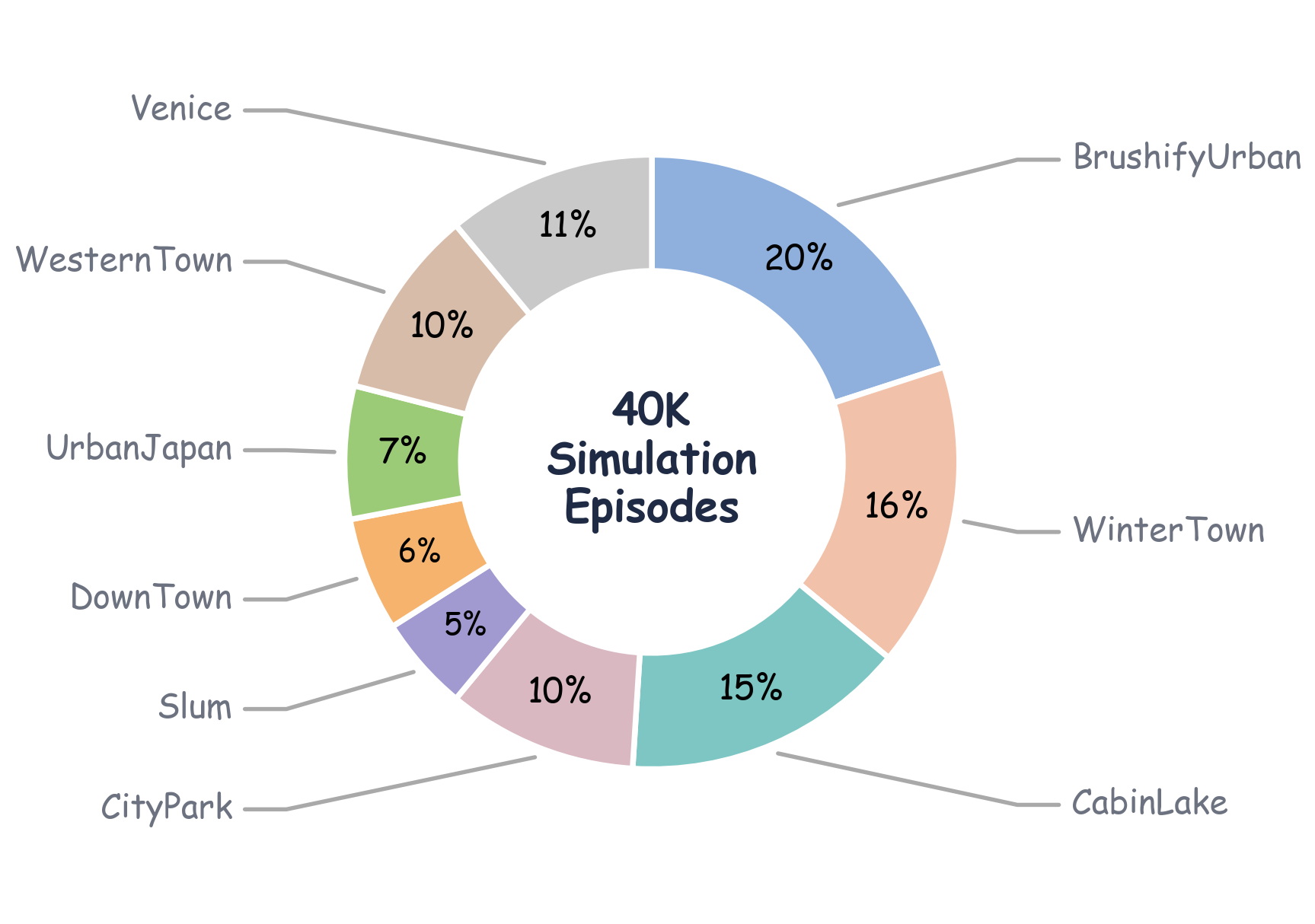}
        \caption{Environment Dist.}
    \end{subfigure}
    \hfill
    \begin{subfigure}[b]{0.24\linewidth}
        \centering
        \includegraphics[width=\linewidth]{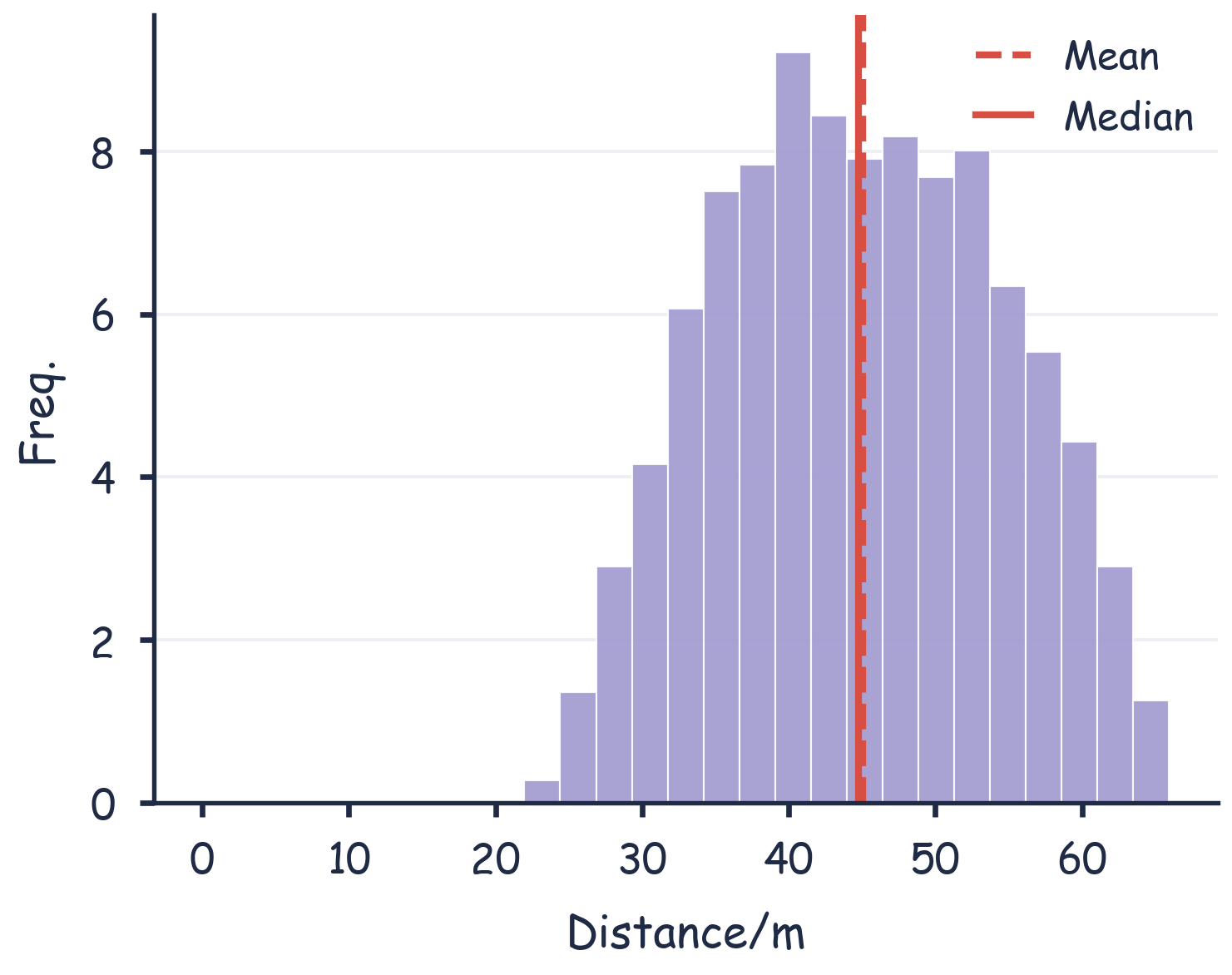}
        \caption{Simulated Traj. Len.}
    \end{subfigure}
    \hfill
    \begin{subfigure}[b]{0.24\linewidth}
        \centering
        \includegraphics[width=\linewidth]{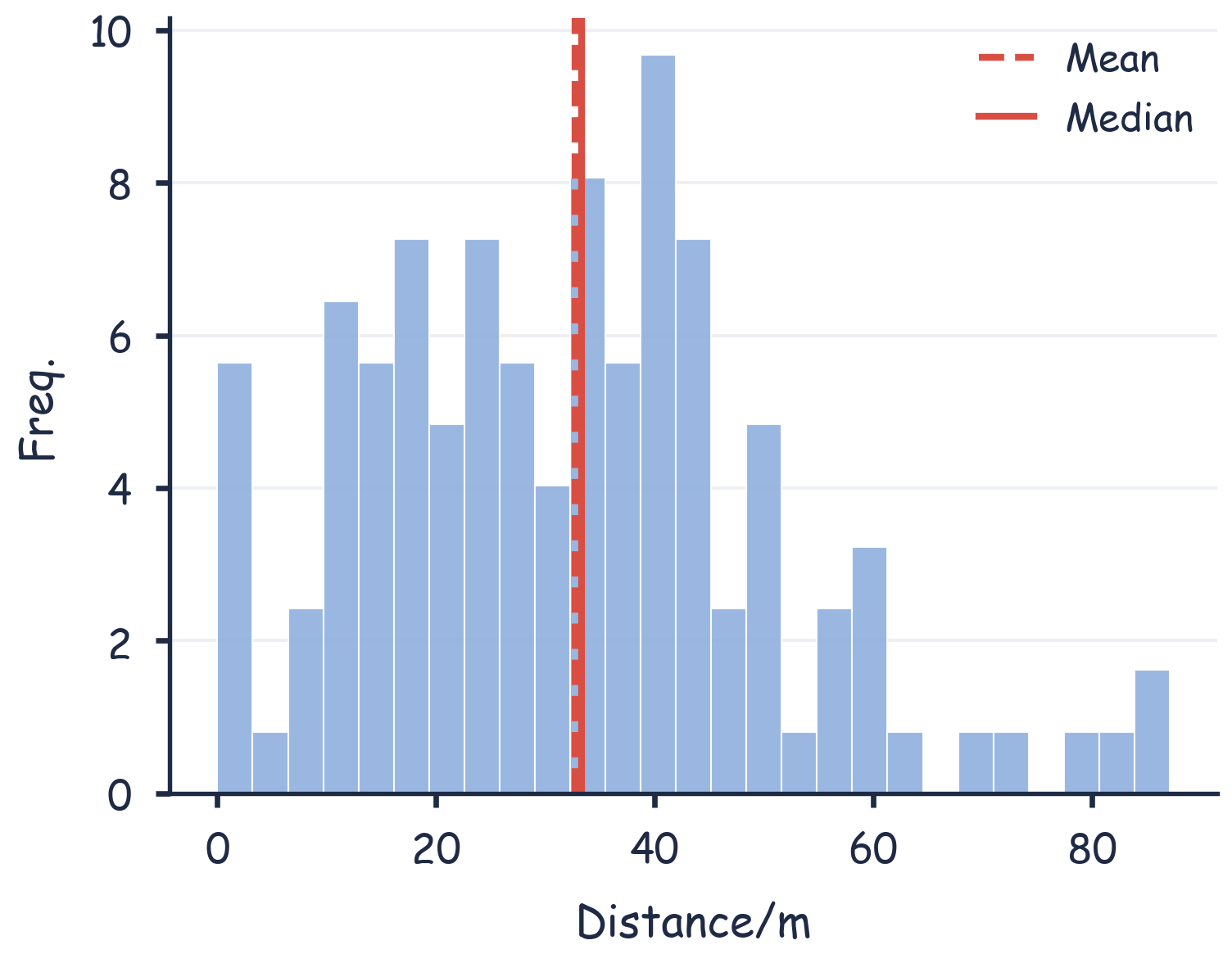}
        \caption{Real-World Traj. Len.}
    \end{subfigure}
    \hfill
    \begin{subfigure}[b]{0.24\linewidth}
        \centering
        \includegraphics[width=\linewidth]{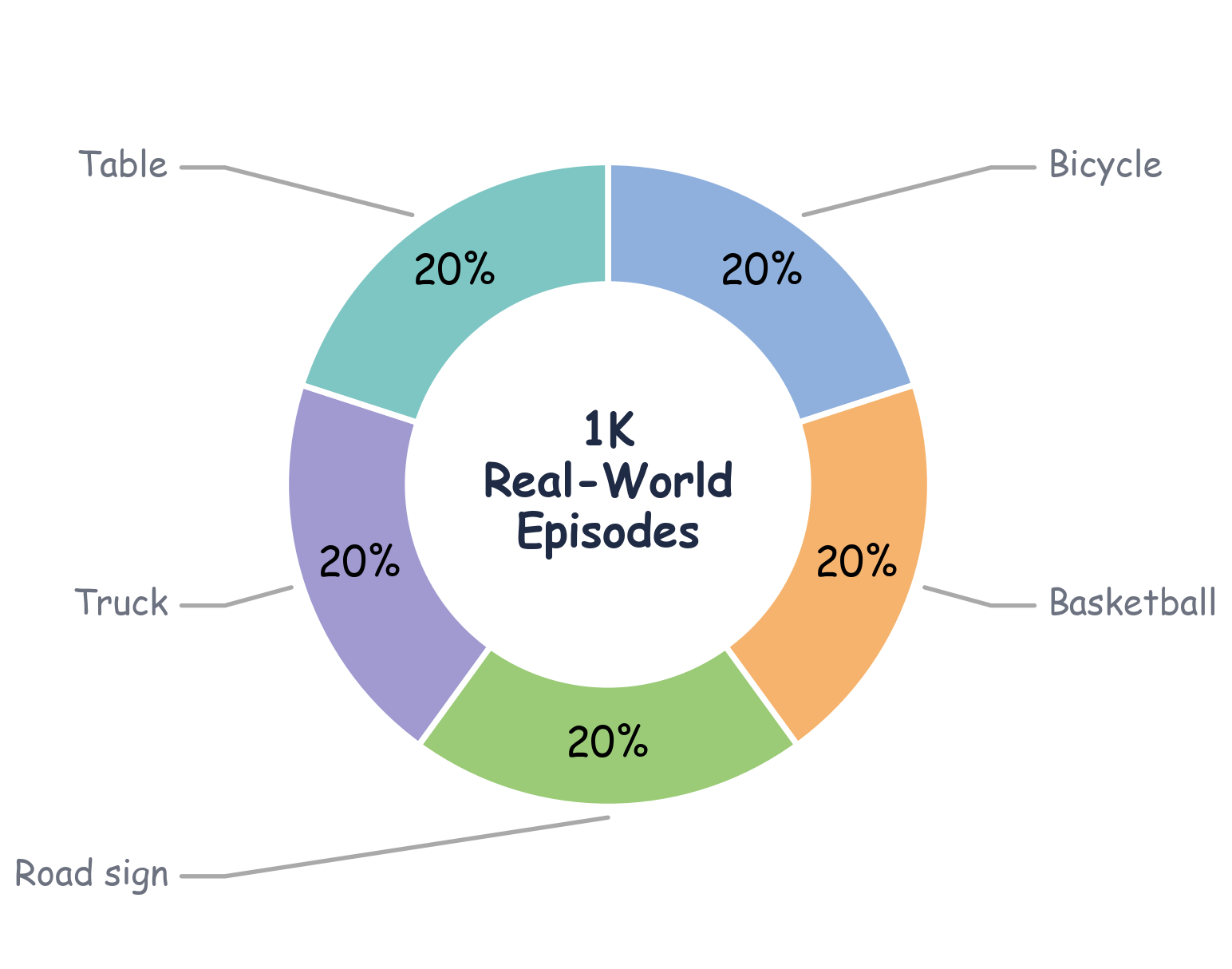}
        \caption{Target Category Dist.}
    \end{subfigure}

    \caption{\textbf{Descriptive statistics of the FG-EQA benchmark.} \textbf{(a)} and \textbf{(d)} Output answer length distributions for the simulated domain-specific VQA and the filtered generic VQA data, respectively (enforcing concise short answers). \textbf{(b)} and \textbf{(c)} Word clouds illustrating semantic focuses. \textbf{(e)} The distribution of the 40K simulated trajectories across 9 high-fidelity digital-twin environments. \textbf{(f)} and \textbf{(g)} Trajectory length (meters) distributions for simulated and real-world perception flights. \textbf{(h)} Distribution of the 1K real-world episodes across 5 distinct physical target categories.}
    \label{fig:data_distribution}
\end{figure}

Figure~\ref{fig:data_distribution} presents a comprehensive descriptive statistical overview of the generated FG-EQA benchmark, validating the diversity and rigor of our dataset construction pipeline from both linguistic and physical dimensions. 

\textbf{Linguistic Properties.} 
Figures~\ref{fig:data_distribution}a and \ref{fig:data_distribution}d depict the answer length distributions for the simulated domain-specific VQA and the filtered generic VQA data, respectively. Both distributions exhibit a sharp peak concentrated within 1-10 words, robustly demonstrating the efficacy of our stringent filtering protocols (e.g., the $\le 32$ characters constraint discussed in Appendix~\ref{app:dataset_qa}). By tightly bounding the response length, we compel the VLM to fundamentally learn localized, evidence-grounded visual reasoning rather than hallucinating lengthy, unconstrained conversational text. Furthermore, the word clouds highlight a deliberate semantic complementarity: the simulated VQA (Figure~\ref{fig:data_distribution}b) is heavily dominated by fine-grained industrial and vehicular attributes (e.g., \textit{``color'', ``wooden'', ``plate'', ``material''}), whereas the integrated generic VQA (Figure~\ref{fig:data_distribution}c) provides broad, open-world semantic coverage (e.g., \textit{``man'', ``woman'', ``person'', ``dog''}). This designed semantic divergence decisively prevents the multimodal backbone from overfitting to monotonic simulation geometries.

\textbf{Spatial and Trajectory Properties.} 
As delineated in Figure~\ref{fig:data_distribution}e, the 40K simulated episodes are well-balanced across 9 complex outdoor environments, with the largest proportion (\textit{BrushifyUrban}) occupying roughly 20\%, thereby guaranteeing rigorous spatial diversity. Correspondingly, Figure~\ref{fig:data_distribution}h verifies that our 1K real-world validation episodes are strictly uniformly distributed (20\% each) across five distinct physical inspection targets. Crucially, the trajectory length distributions in Figures~\ref{fig:data_distribution}f and \ref{fig:data_distribution}g reveal that both simulated and real-world perception flights average between 35 and 45 meters. This spatial displacement indicates that fine-grained active perception is not merely a localized adjustment; rather, it requires continuous spatial maneuvering to bridge the gap between macroscopic target discovery and microscopic evidence acquisition.

\section{Model Details}
\label{app:model}

ScoutVLA is built upon a robust dual-expert architecture designed to intrinsically couple semantic reasoning with continuous spatial control. For visual encoding and language modeling, the system utilizes a SigLIP ViT-So400m/14 encoder~\citep{zhai2023siglip} paired with a PaliGemma~2B multimodal backbone~\citep{steiner2024paligemma2}. The language backbone consists of 18 layers (width 2048, MLP dimension 16384, 8 attention heads) with a shared vocabulary of 257,152 tokens. To mitigate the representation shift introduced by Stage 1 action training, the multimodal hidden states are sequentially processed through a FinalNorm layer and our lightweight LMAdapter before projecting to the vocabulary logits via the Embedder decoder. In parallel, the independent Action Expert is instantiated as a 300M parameter network (width 1024, depth 18, MLP dimension 4096). During practical deployment, the visual inputs are formulated across three image slots to ensure interface compatibility with the OpenPI framework, utilizing the active FPV and initial start frames while zero-filling the redundant slot.

\section{Training and Inference Details}
\label{app:training}

This section provides the implementation specifics required to reproduce the decoupled training paradigm and the closed-loop inference pipeline. The essential hyperparameter configurations are summarized in Table~\ref{tab:training_hyperparams}, omitting standard deep-learning defaults for clarity.

\textbf{Computational Cost and Stage 1 Setup.} All training pipelines are consistently executed across a compute cluster equipped with 8$\times$ NVIDIA A800 GPUs. For Stage 1, both ScoutVLA variants are optimized over 5 full epochs, consuming approximately 120 hours of wall-clock time. The optimization is driven by the AdamW optimizer with a cosine learning rate decay and a 1,000-step linear warmup. We define a continuous action chunk horizon of 10 steps and maintain an Exponential Moving Average (EMA) decay of 0.99 to ensure stability in the continuous flow-matching vector field regression.

\textbf{Stage 2 Knowledge Insulation.} During the Stage 2 joint training phase, which similarly lasts for 5 epochs, we formally enact the proposed Knowledge Insulation mechanism. At the engineering level, we freeze the base PaliGemma network and dynamically toggle gradient masks based on the modality of the current batch. The data loader strictly alternates updates, maintaining a rigid ratio of 3 action-control batches for every 1 language-QA batch. Control steps exclusively optimize the flow-matching parameters, whereas language steps selectively update the rank-16 LoRA modules and the LMAdapter. To accelerate text head convergence without inducing instability, the LMAdapter is assigned a $2.5\times$ learning rate multiplier compared to other semantic parameters.

\textbf{Inference and Deployment.} For physical deployment and simulated evaluation, the control execution phase utilizes Euler solvers for rapid flow-matching denoising. Once the initial target is approached and the visual evidence acquisition is deemed complete by the stop token (threshold $> 0.85$), the agent explicitly transitions to the VQA inference phase. For this terminal phase, ScoutVLA zero-fills the redundant OpenPI image slots and executes autoregressive greedy decoding at a temperature of $0$ to rigorously enforce deterministic and factual short-answer generation.

\begin{table}[t]
    \centering
    \caption{\textbf{ScoutVLA Training Configurations.} Essential hyperparameters for the two-stage decoupled training paradigm.}
    \label{tab:training_hyperparams}
    \begin{tabular}{lc}
        \toprule
        \textbf{Parameter} & \textbf{Value} \\
        \midrule
        Batch Size & 128 \\
        Epochs per Stage & 5 \\
        Peak Learning Rate & $2\times 10^{-5}$ \\
        LoRA Enabled & True (Stage 2) \\
        LoRA Rank & 16 \\
        Action Horizon Steps & 10 \\
        Training Time & $\sim$120 hours per stage \\
        Hardware & 8 $\times$ NVIDIA A800 GPU \\
        \bottomrule
    \end{tabular}
\end{table}

\section{Evaluation Protocol}
\label{app:evaluation}

\subsection{Metrics Definition}

To rigorously decouple and evaluate both the spatial maneuvering efficiency and the complex multimodal reasoning capabilities of the EQA agents, we establish a comprehensive evaluation suite comprising six specialized metrics. For a total of $N$ evaluation episodes, we formally define the following components: Let $d_i \in \{0,1\}$ denote a binary indicator of whether the underlying target object is successfully localized within the bounding frustum of the final frame. Let $s_i \in [1,5]$ represent the continuous QA quality score, derived as the arithmetic mean from our independent LLM proxy panel. We define $c_i = \mathbb{I}[s_i > 4]$ as the categorical indicator for discrete QA correctness (where answers scoring strictly above 4 are considered factually correct). Furthermore, let $\delta_i$ represent the Euclidean distance (in meters) from the UAV's terminal coordinate to the target centroid, and $\ell_i$ denote the accumulative trajectory distance navigated during the episode. Based on these primitives, we compute the following three categories of metrics:

\textbf{1. Spatial Navigation Metrics.} These metrics evaluate the pure flight control and target-seeking bounds independent of language generation:
\begin{align}
    \mathrm{TDR} &= \frac{100}{N}\sum_{i=1}^{N} d_i \quad &\text{(\textbf{Target Detection Rate})} \\
    \mathrm{DT} &= \frac{1}{N}\sum_{i=1}^{N} \delta_i \quad &\text{(\textbf{Distance to Target})} \\
    \mathrm{ATL} &= \frac{1}{N}\sum_{i=1}^{N} \ell_i \quad &\text{(\textbf{Average Trajectory Length})}
\end{align}
TDR measures the macroscopic success of bringing the target into the terminal Field of View. For passive QA baselines that inherently lack mobility, these navigation-dependent metrics are fundamentally not applicable.

\textbf{2. Multimodal Reasoning Metrics.} These metrics evaluate the raw textual reasoning accuracy against the visual evidence provided by the terminal view:
\begin{align}
    \mathrm{QAS} &= \frac{1}{N}\sum_{i=1}^{N} s_i \quad &\text{(\textbf{QA Score})} \\
    \mathrm{QAC} &= \frac{100}{N}\sum_{i=1}^{N} c_i \quad &\text{(\textbf{QA Correctness})}
\end{align}
QAS provides a granular assessment of answer completeness and formatting, while QAC serves as a harsh binary threshold for factual accuracy.

\textbf{3. Joint Success Metric.} To capture the overarching goal of active perception, we introduce the most rigorous EQA metric:
\begin{align}
    \mathrm{SSR} &= \frac{100}{N}\sum_{i=1}^{N} d_i c_i \quad &\text{(\textbf{Strict Success Rate})}
\end{align}
SSR enforces a strict logical ``AND'' condition. An episode is deemed successful \emph{if and only if} the autonomous flight successfully captures the target within the terminal view ($d_i=1$) \emph{and} the downstream language head subsequently generates the functionally correct answer ($c_i=1$). This joint metric definitively reflects whether the entire semantic-to-physical loop is solved.

\subsection{Baseline Evaluation Protocol}
\label{app:baseline}

Given the fundamental architectural and functional discrepancies among the evaluated models, mapping them to a unified evaluation framework requires standardizing the pipelines. We systematically categorize all baselines into three distinct families, explicitly outlining the algorithmic characteristics and the specific evaluation procedures for each:

\begin{itemize}
    \item \textbf{General VLM and Passive QA Baselines:}
    This category represents static language or vision-language models that do not interact with the physical environment. They act entirely as passive observers without the capability to generate continuous actions.
    \begin{itemize}
        \item \textbf{Blind QA / Language-Only:} These models explicitly receive the textual question while the visual input is completely zeroed out. They serve as a robust ablation probe to detect underlying language biases within the FG-EQA benchmark, fundamentally verifying that the questions cannot be guessed via ungrounded common sense.
        \item \textbf{Initial-View VLMs (e.g., Qwen2-VL~\citep{wang2024qwen2vl}, Qwen3-VL~\citep{bai2025qwen3vl}):} Representing state-of-the-art generalist multimodal reasoning, these models receive exactly the sub-optimal initial FPV image alongside the complex query. Their zero-shot performance directly defines the theoretical upper bound of static visual reasoning when fine-grained active mobility is absent.
    \end{itemize}

    \item \textbf{Navigation-Only Baselines:}
    These are contemporary, dedicated aerial embodied policies. While they output functional flight navigation sequences (either continuous or discrete), they fundamentally lack a natural-language generation head capable of executing EQA.
    \begin{itemize}
        \item \textbf{OpenFly~\citep{gao2025openfly}:} An autoregressive 7B policy structured for long-range, point-to-point macroscopic routing via discrete waypoint generation. It fundamentally prioritizes destination arrival over semantic viewpoint alignment.
        \item \textbf{TravelUAV~\citep{wang2024traveluav}:} An aerial framework operating via continuous 6-DoF control. However, it relies heavily on multi-view geometric camera arrays and is engineered toward collision avoidance rather than monocular active visual search. 
        \item \textbf{UAV-Flow~\citep{wang2025uavflow}:} Employs an advanced diffusion/flow-based head to regress continuous low-level flight actions. Crucially, despite possessing robust spatial maneuverability for short-horizon path following, it lacks any multimodal decoder to emit textual answers.
    \end{itemize}
    \textit{Standardized Evaluation:} For all navigation-only baselines, we first uniquely unroll their predicted active trajectories in the simulator until their respective internal sequence- termination conditions are met. We then extract their final captured FPV image and uniformly query an external, frozen Qwen3-VL with the original text question. This paradigm rigorously decouples their spatial search capability from textual decoding constraints.

    \item \textbf{Embodied EQA Baselines:}
    These agents are comprehensively designed for the full EQA task. We evaluate both modular navigate-then-answer frameworks and our end-to-end approach, which output both the sequential spatial maneuvers and the terminal textual answers.
    \begin{itemize}
        \item \textbf{CityEQA~\citep{zhao2025cityeqa}:} A prominent outdoor embodied framework engineered for macroscopic target search. It typically issues a stop action the moment the generic target enters the FOV, inherently lacking the architecture for subsequent fine-grained micro-viewpoint refinement.
        \item \textbf{CoV~\citep{zhao2026cov}:} A state-of-the-art indoor EQA paradigm. We introduce it to explicitly evaluate whether visual-navigation strategies tailored for heavily constrained indoor spatial geometries can successfully adapt to sparse, unconstrained outdoor aerial environments.
        \item \textbf{ScoutVLA (Ours):} Employs a dual-expert integration mechanism driving flow matching generation. It optimally addresses active perception by seamlessly transitioning from macroscopic target detection directly into 5-DoF continuous micro-refinement trajectories, culminating in factual short-answer synthesis.
    \end{itemize}
\end{itemize}

\textbf{Terminal-View Answerability Analysis.} 
As discussed in Section~\ref{subsec:quality_analysis}, evaluating the raw information gain of an active trajectory requires isolating the view generation behavior from the model's inherent language capacity. Thus, during this specific sub-evaluation, all terminal visual endpoints provided by the active embodied agents are frozen. We subsequently query an external panel of state-of-the-art VQA models acting uniformly as answer generators. Their textual outputs are identically scored using the human-aligned LLM judge protocol detailed below.

\subsection{LLM Judge Prompt Template}

To rigorously minimize intra-model heuristic bias and mitigate unpredictable subjective variance during answer validation, our evaluation protocol employs a diverse proxy panel consisting of three frontier Large Language Models (Qwen3-VL, Kimi-K2.6, and GLM-5.1) acting as independent, blinded answer judges. Following the robust benchmarking practices established in OpenEQA~\citep{majumdar2024openeqa} and CityEQA~\citep{zhao2025cityeqa}, we aggregate these independent assessments and report the exact arithmetic mean as the final score. 

\begin{figure}[h]
    \centering
    \includegraphics[width=0.9\linewidth]{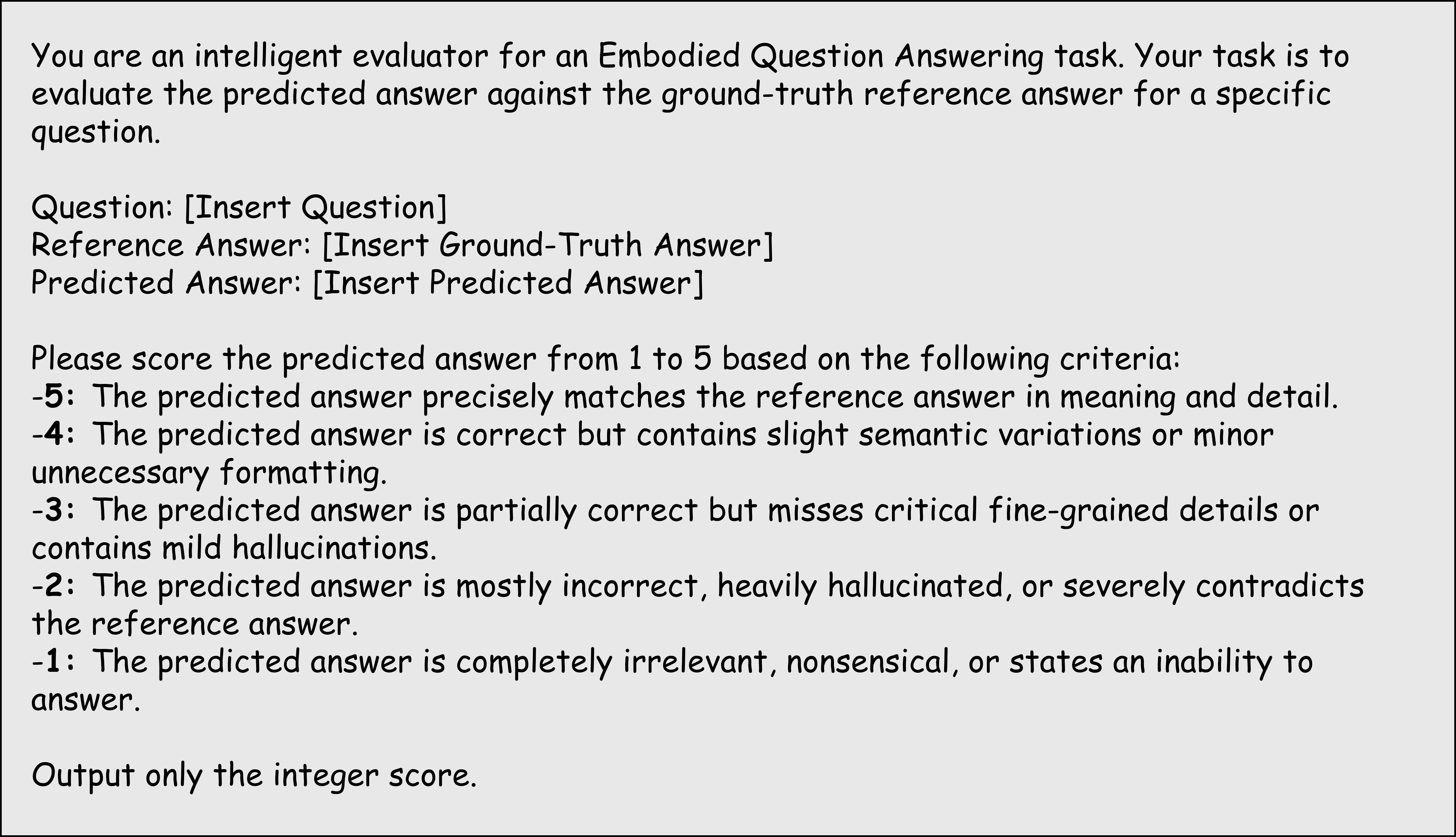}
    \caption{\textbf{Standardized LLM Judge Evaluation Prompt.} The deterministic scoring template utilized by the proxy panel. This prompt strictly forces the LLM evaluators to cross-reference the agent's predicted answer against the ground truth based on a granular, 5-tier semantic alignment mechanism, preventing uncontrolled text-generation variability.}
    \label{fig:f3_prompt}
\end{figure}

During evaluation, each LLM judge receives a unified textual prompt containing the contextual semantic question, the ground-truth reference answer retrieved from our dataset, and the unverified generated answer produced by the embodied agent. The formalized evaluation template is designed to strictly constrain the model's output to an integer scale from 1 (completely irrelevant or hallucinatory) to 5 (precise, meaning-aligned match). The complete text configuration of this unified 1--5 scoring prompt is visually presented in Figure~\ref{fig:f3_prompt}.

\section{Experimental Results and Deployment}
\label{app:experiments_deployment}

\subsection{Real-World Deployment Architecture}
\label{app:deployment_setup}

To deploy ScoutVLA on a real UAV, we designed a highly decoupled, hardware-in-the-loop deployment architecture. While deploying a multi-billion parameter VLA model directly onboard is theoretically feasible via heavy advanced edge-compute modules, doing so imposes exceptionally high hardware requirements on standard commercial off-the-shelf UAVs. Specifically, integrating massive local GPU accelerators severely compromises the UAV's payload budget, drastically drains battery life, and escalates platform scaling costs.

\begin{figure}[t]
    \centering
    \includegraphics[width=0.6\linewidth]{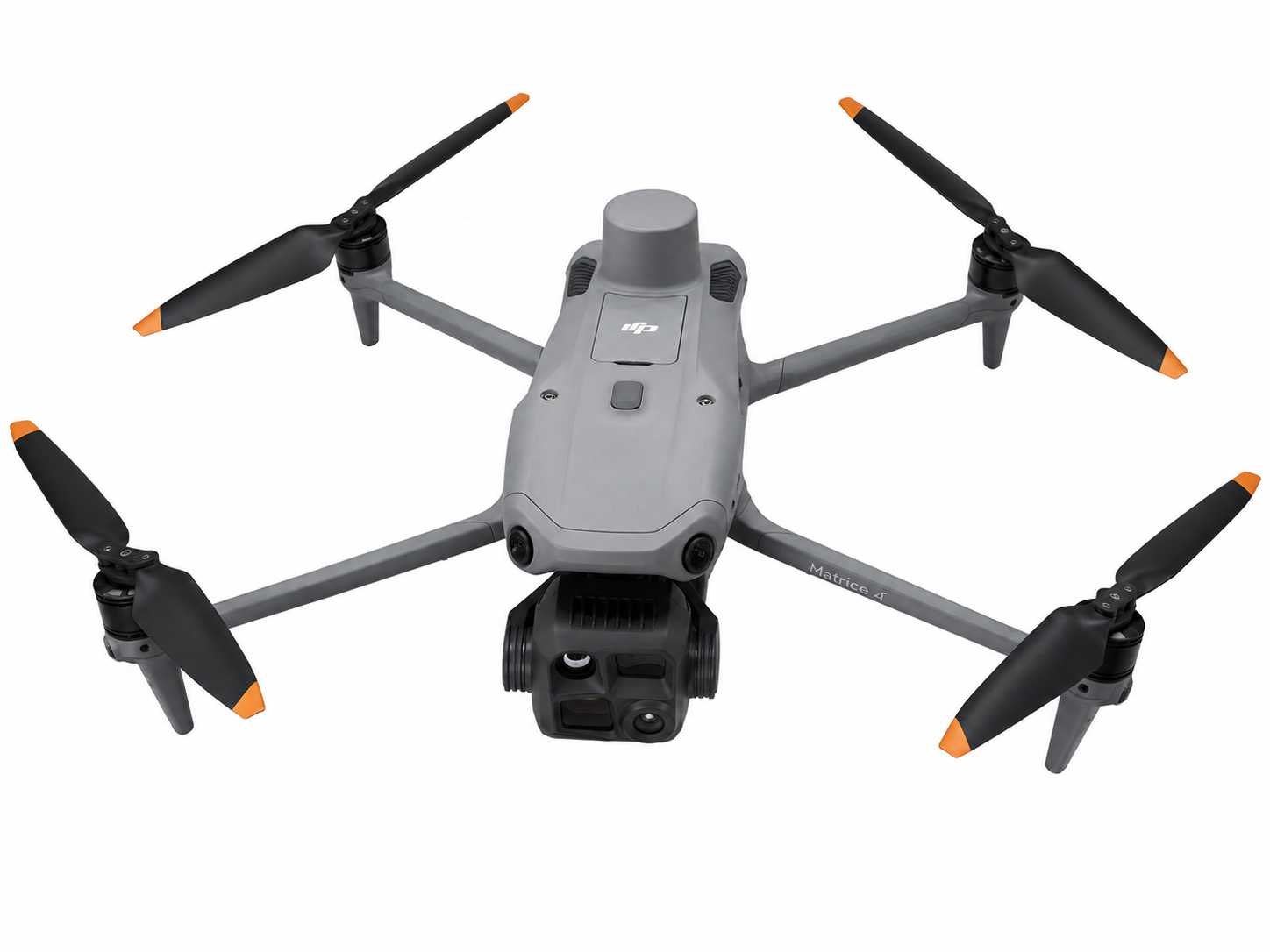} 
    \caption{\textbf{Physical Hardware Setup.} The commercial off-the-shelf DJI UAV utilized during our real-world physical deployment. By establishing a decoupled hardware-in-the-loop bridge between the UAV's remote controller and the high-performance local ground station, ScoutVLA actively governs the continuous flight maneuvers relying exclusively on the drone's standard onboard monocular RGB camera.}
    \label{fig:uav_hardware}
\end{figure}

Consequently, we shift the heavy computational burden by deploying the frozen ScoutVLA policy exclusively on a high-performance local edge ground station (equipped with an NVIDIA RTX 5090 GPU). The physical commercial platform strictly utilized for all unconstrained outdoor deployments is explicitly illustrated in Figure~\ref{fig:uav_hardware}. During live physical inspection tasks, an uncompromised closed-loop perception-action cycle is established: The UAV's onboard monocular camera captures real-time visual perceptions and transmits them over a low-latency wireless transmission downlink directly to the handheld Remote Controller (RC). The RC, tethered to the local processing machine via a high-bandwidth wired protocol, continuously feeds these sequential FPV frames into the local neural network. ScoutVLA subsequently executes rapid flow-matching inference to predict the necessary 5-DoF continuous spatial adjustments. These generated action increments are instantaneously routed back to the RC, which translates them into concrete radio-frequency control signals to actively actuate the UAV's flight controller. 

\subsection{Supplementary Experimental Analysis}
\label{app:supplementary_analysis}

Due to strict page limitations in the main manuscript, several critical mechanistic insights derived from our experiments (Section~\ref{sec:experiments}) are comprehensively elaborated here.

\textbf{Analysis of SOTA Baselines (ref. Section~\ref{subsec:sota}).} 
The performance gap in SSR between ScoutVLA and general VLMs empirically confirms that raw multimodal reasoning cannot substitute for active physical mobility. While Qwen2-VL localized the target in 57.41\% of simulated episodes, its terminal viewpoints remained statically bound to the initial sub-optimal poses, keeping the critical evidence heavily occluded and strictly capping its final QAC. Conversely, while exclusively navigation-driven baselines like OpenFly and TravelUAV possess the required mobility, their autoregressive and explicitly discrete action formulation restrict them to macroscopic obstacle avoidance or coarse point-to-point waypoint routing. This fundamental architectural mismatch actively prevents them from executing the micro-scale, millimeter-level 5-DoF continuous refinements required to reliably expose highly angle-dependent visual cues. By natively unifying continuous flow-matching with linguistic multimodal reasoning, ScoutVLA directly bypasses this systemic bottleneck.

\textbf{Analysis of Decoupled Training (ref. Section~\ref{subsec:ablation}).} 
The targeted ablation studies in the main text highlight the risk of catastrophic forgetting during unconstrained multimodal joint learning. When we systematically ablated the Knowledge Insulation mechanism, both navigation competence and reasoning accuracy crashed simultaneously to near-zero levels. This catastrophic degradation occurs because predicting real-valued, high-frequency continuous action vectors inherently generates massive gradient magnitudes. If left uninsulated, these action gradients entirely overwhelm and destructively overwrite the delicate, pre-trained parameters of the autoregressive language head, erasing the model's textual decoding capability. By rigidly freezing the core semantic backbone and strategically bridging the discrete-continuous modalities exclusively via parameter-efficient LoRA adapters and our proposed LMAdapter, the two-stage decoupled recipe forces the continuous control head to rapidly adapt to the stable visual representations, deliberately preventing the structural mutual destruction of the shared multimodal weights.

\textbf{Underlying Logic of Terminal-View Answerability (ref. Section~\ref{subsec:quality_analysis}).} 
A notable finding from the main text is that when utilizing frozen terminal images, external agnostic VLMs achieve drastically higher QA scores when evaluating ScoutVLA's generated endpoints compared to endpoints from any baseline policy. This validates a profound assertion: the intrinsic value of active perception is completely independent of the agent's internal language capacity. ScoutVLA geometrically restructures the environment to optimize the subsequent semantic information gain. Consequently, by successfully translating abstract visual linguistic queries into 5-DoF active spatial alignment, the architecture unequivocally transforms a previously unanswerable scenario into a highly deterministic and factually localized visual configuration, thereby improving the performance bound for downstream reasoning models.

\textbf{Quantitative Results in Physical Deployment via Sim-to-Real Transfer.}
To comprehensively validate the Sim-to-Real transferability of our architecture, we adopt a sequential fine-tuning paradigm. Rather than training blindly from scratch, we initialize the process by utilizing the ScoutVLA-Pi0 and ScoutVLA-Pi05 checkpoints that have been fully pre-trained on the massive 40K simulation trajectories. These robust foundational models—already possessing rich spatial intelligence—are subsequently subjected to a two-phase real-world fine-tuning process. First, the independent action expert is fine-tuned exclusively employing the targeted 1K real-world perception trajectories, allowing it to adapt to real aerodynamic dynamics and sensor noise. Following this, the multimodal language components are selectively aligned utilizing only authentic real-world generic VQA data (strictly excluding all AirSim-simulated visual-language pairs). In doing so, we efficiently bridge the visual and physical domain gaps.

The resultant models are then seamlessly deployed onto our physical framework. As reported in Table~\ref{tab:real_world_metrics}, we execute 50 closed-loop physical inspection flights (10 independent trials strictly conducted across 5 distinct target categories). Remarkably, the fine-tuned ScoutVLA-Pi05 accomplishes a SSR of 68.30\% under unconstrained outdoor conditions. This robust empirical performance definitively confirms that our simulated pre-training acts as a potent structural anchor, effortlessly unlocking high-fidelity real-world active perception via minimal, targeted physical fine-tuning.

\begin{table}[h]
    \centering
    \caption{\textbf{Quantitative Sim-to-Real Results in Physical Deployment.} Evaluation of the fine-tuned ScoutVLA models across 50 physical inspection flights (10 independent trials per target category). The results systematically prove the efficacy of our Sim-to-Real transfer pipeline.}
    \label{tab:real_world_metrics}
    \resizebox{\linewidth}{!}{%
    \begin{tabular}{lcccccc}
        \toprule
        \textbf{Method} & \textbf{TDR (\%)} $\uparrow$ & \textbf{DT (m)} $\downarrow$ & \textbf{ATL (m)} & \textbf{QAS} $\uparrow$ & \textbf{QAC (\%)} $\uparrow$ & \textbf{SSR (\%)} $\uparrow$ \\
        \midrule
        ScoutVLA-Pi0 (Real) & 75.30 & \textbf{5.1} & \textbf{22.4} & 4.12 & 65.40 & 60.10 \\
        ScoutVLA-Pi05 (Real)& \textbf{81.50} & 5.8 & 24.1 & \textbf{4.28} & \textbf{71.20} & \textbf{68.30} \\
        \bottomrule
    \end{tabular}}
\end{table}

\subsection{Qualitative Success Cases}
\label{app:success_cases}

\begin{figure}[t]
    \centering
    \includegraphics[width=\linewidth]{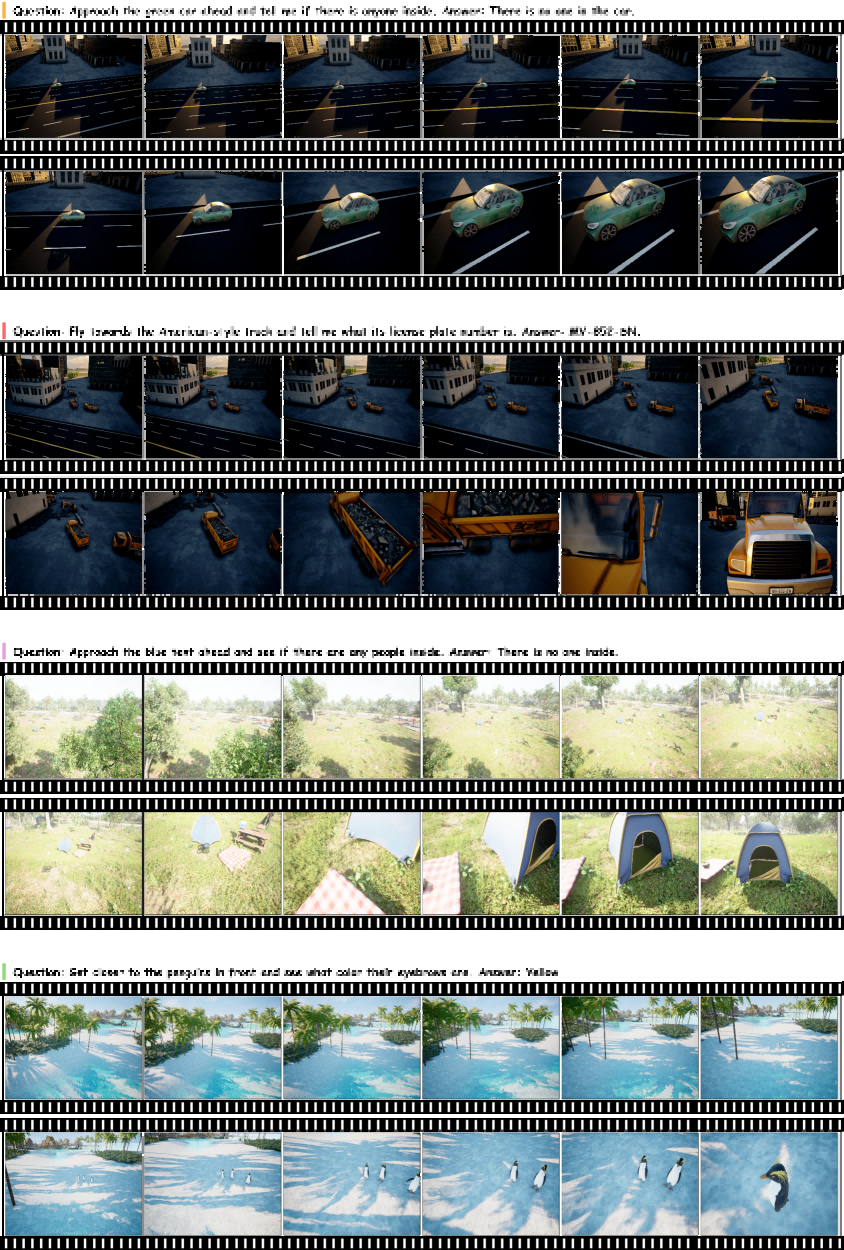}
    \caption{\textbf{Sequential Success Cases in FG-EQA Simulation.} The continuous frame-by-frame sequences demonstrate ScoutVLA actively refining its viewpoint across various scenes (e.g., nocturnal streets, industrial zones, beaches). It successfully transitions from a visually ambiguous initial discovery (top rows) to a microscopic, detailed terminal observation (bottom rows).}
    \label{fig:success_sim}
\end{figure}

\textbf{Simulation Scenarios.}
In highly cluttered or visually challenging digital-twin environments, macroscopic routing frequently terminates at sub-optimal distances. As illustrated by the sequential film-strip frames in Figure~\ref{fig:success_sim}, ScoutVLA explicitly overcomes this premature termination across highly diverse topologies. In these demanding test cases—ranging from a nocturnal urban street and an industrial construction zone to a grassy campsite and a coastal beach with penguins—the initial frames (top rows of each sub-panel) capture only general geometric outlines or heavily shadowed profiles. Utilizing its flow-matching action expert, ScoutVLA systematically generates a continuous spatial refinement trajectory. The drone smoothly descends and continuously modifies its Cartesian coordinates and camera pitch, tracking the corresponding entities (e.g., the green vehicle or the penguins) across the terrain. The resulting terminal frames (bottom rows) showcase improved visual clarity, firmly bringing the previously illegible microscopic evidence into the center of the FPV, directly enabling accurate EQA.

\begin{figure}[t]
    \centering
    \includegraphics[width=\linewidth]{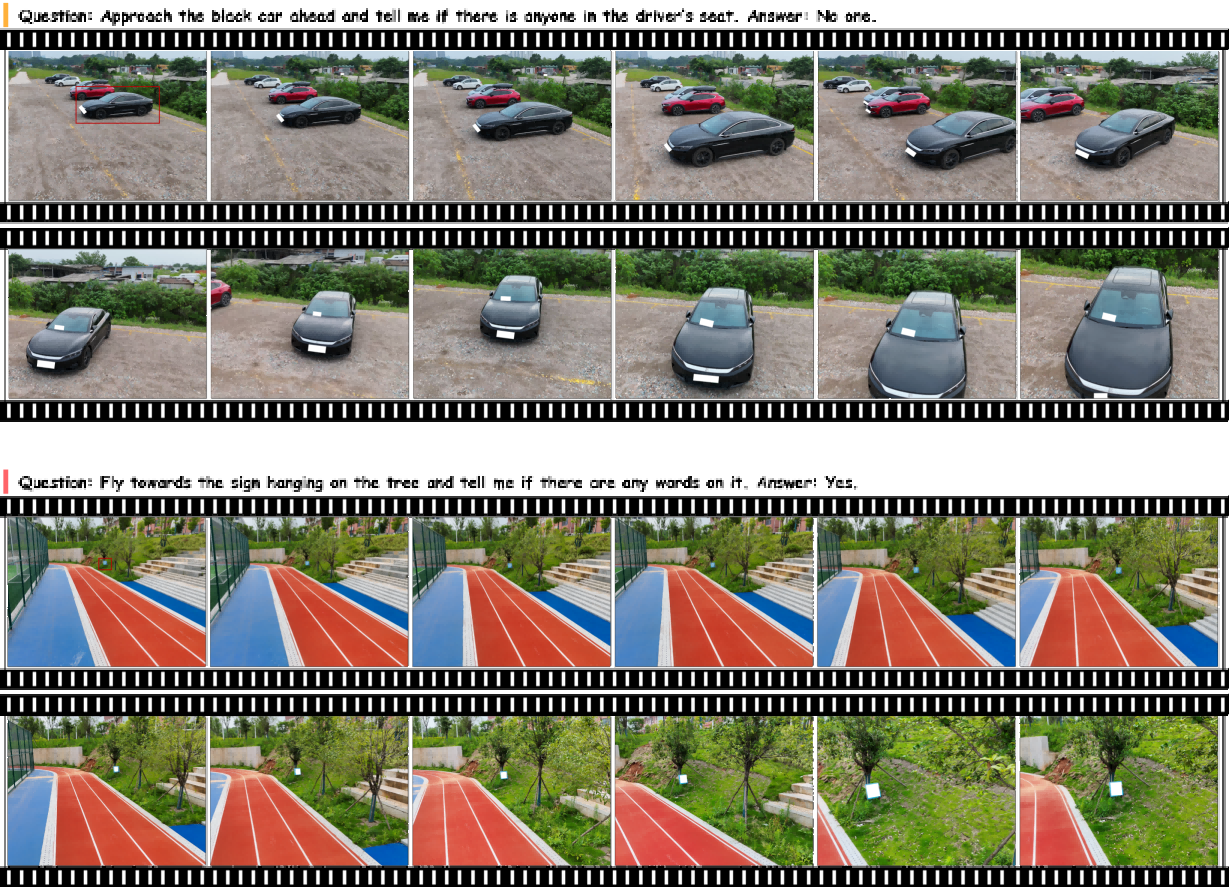}
    \caption{\textbf{Sequential Success Cases in Real-World Deployment.} Under authentic physical interference, ScoutVLA smoothly transitions from high-altitude macroscopic tracking (top rows) to precise, ground-level micro-inspections (bottom rows) around real vehicles and athletic tracks, securing highly detailed close-ups against real-world terrain textures.}
    \label{fig:success_real}
\end{figure}

\textbf{Real-World Scenarios.}
Operating under actual physical constraints introduces dynamic lighting shifts and aerodynamic variability largely absent in simulation. As depicted in the sequential frames of Figure~\ref{fig:success_real}, ScoutVLA maintains exceptional target-locking capabilities under physical deployment. For instance, whether inspecting a black sedan in an unconstrained parking lot or tracking along a vibrant athletic track to locate specific ground-level signage, ScoutVLA initiates a highly stable, continuous spiraling approach. By dynamically adjusting the gimbal pitch in sync with its forward translation, the policy actively counters environmental reflections and physical distance constraints. The terminal FPV frames confirm that ScoutVLA successfully pinpoints the exact inspection angle, providing a highly localized and illuminated close-up (e.g., exposing the front fascia of the vehicle unambiguously), leading directly to successful VQA resolution.

\subsection{Failure Mode Analysis}
\label{app:failure_modes}

\begin{figure}[t]
    \centering
    \includegraphics[width=\linewidth]{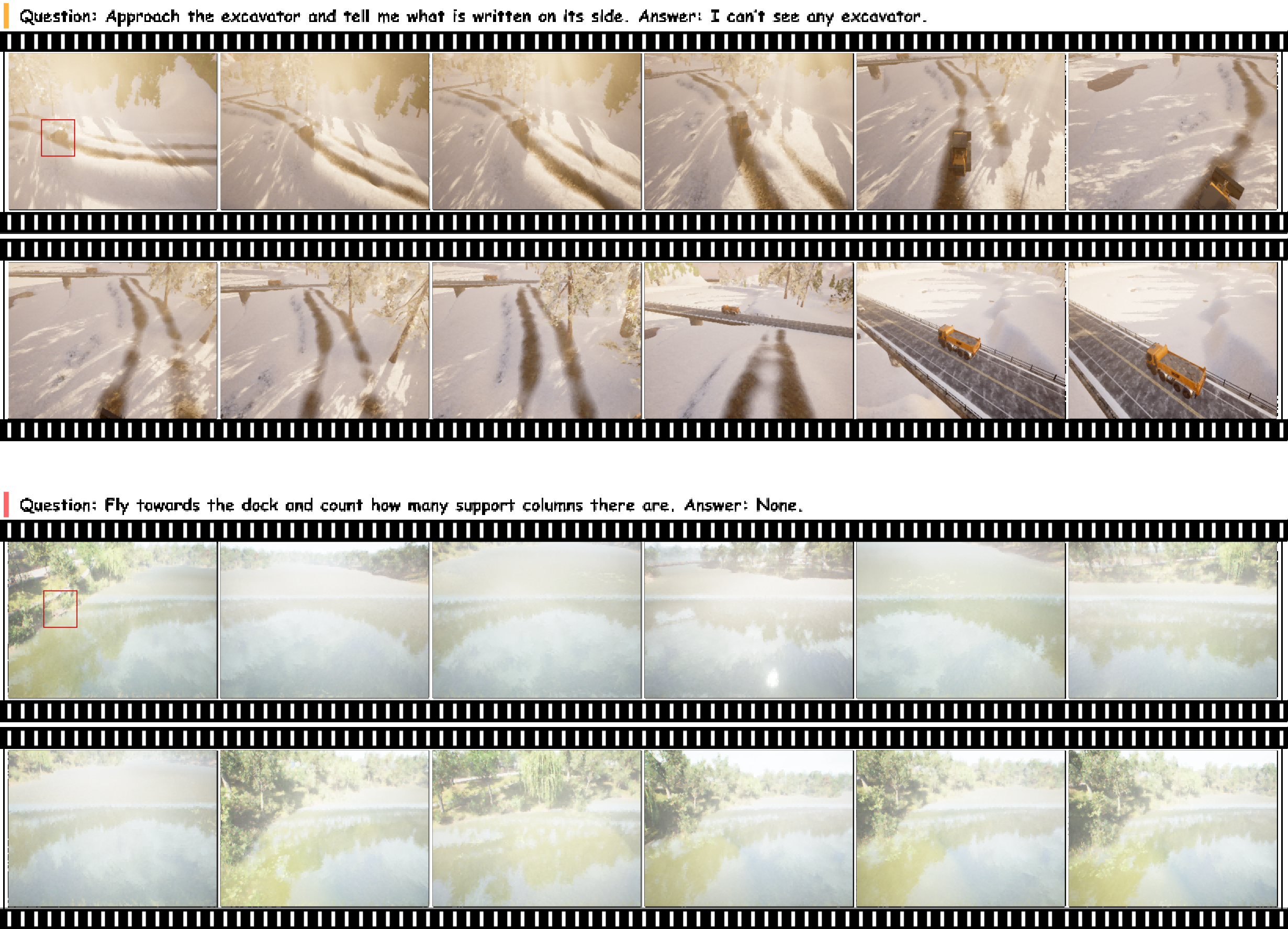}
    \caption{\textbf{Failure Analysis in Simulation.} Top: Severe texture loss in snow causes semantic target hijacking, leading the UAV away from the excavator toward a distractor yellow truck. Bottom: Overwhelming specular reflection on a lake destroys temporal feature consistency, trapping the UAV in a spinning continuous-action local minimum.}
    \label{fig:fail_sim}
\end{figure}

\textbf{Simulation Scenarios.}
Through extensive trajectory logging, we identified two cardinal failure modes in digital-twin environments, exposing fundamental structural vulnerabilities in vision-action alignment (Figure~\ref{fig:fail_sim}):
\begin{itemize}
    \item \textbf{Target Hijacking via Semantic Feature Drift:} In the top snowy environment, the initial target was exclusively a distant excavator. However, the extreme albedo of the expansive snow effectively washes out high-frequency visual textures. Consequently, ScoutVLA's visual encoder degrades to relying on coarse, low-frequency chromatic signals (e.g., the bright yellow hue). This visual ambiguity causes a nearby construction truck of a similar color to act as an overwhelming visual distractor. The continuous policy undergoes ``semantic drift,'' erroneously locking onto the distractor truck and ultimately misguiding the trajectory away from the true target.
    \item \textbf{Perceptual Collapse via Specular Reflection:} In the bottom lake scenario, the designated target was a wooden pier. As the UAV approaches, severe sun glare and unpredictable specular reflections emerge on the water surface. Since the flow-matching action space inherently relies on temporal feature consistency, these volatile radiometric artifacts completely destroy stable optical tracking. The visual-action vector field degrades into conflicting gradients, trapping the drone in a severe spatial ``local minimum,'' manifesting behaviorally as the UAV helplessly spinning in place until sequence termination. 
\end{itemize}

\textbf{Real-World Scenarios.}

\begin{figure}[t]
    \centering
    \includegraphics[width=\linewidth]{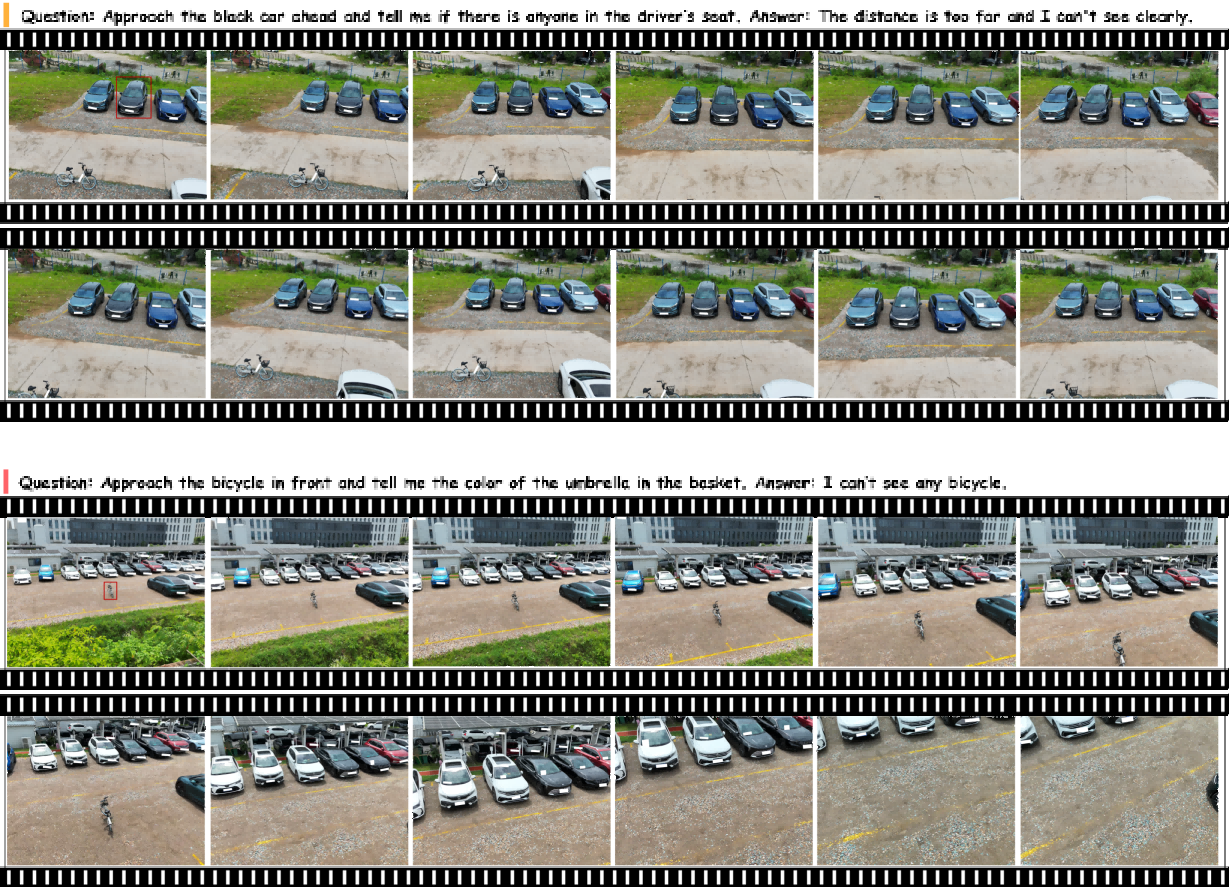}
    \caption{\textbf{Failure Analysis in Real-World Deployment.} Top: The dense presence of homogeneous distractor vehicles paralyzes the continuous action field, causing the drone to stall distantly. Bottom: A kinematic overshoot pushes the small target (bicycle) out of the FOV. Lacking structured geometric memory (object permanence), the agent aimlessly searches empty concrete and irreversibly loses the target.}
    \label{fig:fail_real}
\end{figure}

Extensive telemetry from physical deployments further exposes the intrinsic limitations of current single-view VLA paradigms. As shown in Figure~\ref{fig:fail_real}, real-world outdoor failures frequently manifest through two specific behavioral patterns:
\begin{itemize}
    \item \textbf{Action Paralysis via Homogeneous Distractors:} In the upper sequence, the designated target (a black sedan) is positioned among multiple identical vehicles. As the UAV initiates its approach, the spatial vectors predicted by the flow-matching head begin to conflict. Mechanistically, when multiple identical visual attractors saturate the latent space, the vector field geometry flattens. Unable to isolate unique, discriminative features of the target car, the regression output collapses into near-zero translational increments. Behaviorally, the drone stalls ("hesitates") indefinitely at a sub-optimal distance, exposing a critical deficiency in current multimodal experts: the inability to maintain robust, fine-grained object tracking lock amidst dense homogeneity.
    \item \textbf{Kinematic Overshoot and Visual Amnesia:} In the lower sequence, the UAV is tasked with identifying a small bicycle. While descending aggressively, the drone slightly overshoots its kinematic trajectory, causing the tiny target to slip completely out of the limited FOV. Once the bicycle exits the frame, the agent's spatial awareness abruptly terminates. Despite blindly panning the camera in an attempt to recover, it inevitably fails. This failure mode indicates that ScoutVLA—like most autoregressive or purely reactive contemporary visual-motor policies—lacks ``Object Permanence'' (the geometric memory of objects outside the immediate optical frustum). Without a temporal, world-grounded 3D memory bank, the agent struggles to execute spatial backtracking once the visual anchor is momentarily lost.
\end{itemize}

